\documentclass[Afour,sageh,times]{sagej}

\usepackage{natbib}
\graphicspath{{figures/}{./}}
\usepackage[svgnames]{xcolor}
\usepackage[bookmarks=true, colorlinks=true, allcolors=Black, citecolor=Navy]{hyperref}
\usepackage{relsize}
\usepackage{amsfonts, amsmath, amssymb}

\usepackage{mathnotation}
\usepackage{zref-clever}
\renewcommand{\bodyvel}{\fibercirc}
\renewcommand{\bodyforce}{\groupcov{\worldforce}}
\usepackage{siunitx}

\usepackage{eqparbox}

\usepackage{amsthm}
\usepackage{thmtools}
\declaretheorem[style=definition,qed=$\square$]{example}

\renewcommand{\qed}{\unskip\nobreak\,\qedsymbol}
\declaretheorem[style=definition]{definition}

\declaretheorem{lemma}
\declaretheorem{corollary}

\usepackage{etoolbox}

\let\theparentequation\theequation
\patchcmd{\theparentequation}{equation}{parentequation}{}{}

\renewenvironment{subequations}[1][]{%
\refstepcounter{equation}%
\setcounter{parentequation}{\value{equation}}%
\setcounter{equation}{0}%
\def\theequation{\theparentequation\alph{equation}}%
\let\parentlabel\label%
\ifx\\#1\\\relax\else\label{#1}\fi%
\ignorespaces
}{%
\setcounter{equation}{\value{parentequation}}%
\ignorespacesafterend
}

\newcommand*{\nextParentEquation}[1][]{%
\refstepcounter{parentequation}%
\setcounter{equation}{0}%
\ifx\\#1\\\relax\else\parentlabel{#1}\fi%
}

\newcommand{\dragmetric}{\metric}
\newcommand{\Findragmetric}{\Finmetric}
\newcommand{\fricco}{c}

\begin{document}

\title{Asymmetric Friction in Geometric Locomotion}
\author{Ross L. Hatton, Yousef Salaman, and Shai Revzen}

\affiliation{Ross L. Hatton and Yousef Salaman are with the Robotics program at Oregon State University. Shai Revzen is with the Department of Electrical Engineering and Computer Science at the University of Michigan.}

\corrauth{Ross L. Hatton, Oregon State University}

\email{Ross.Hatton@oregonstate.edu}

\newcommand{\slot}{\square}
\begin{abstract}

Geometric mechanics models of locomotion have provided insight into how robots and animals use environmental interactions to convert internal shape changes into displacement through the world, encoding this relationship in a ``motility map''.
A key class of such motility maps arises from (possibly anisotropic) linear drag acting on the system's individual body parts, formally described via Riemannian metrics on the motions of the system's individual body parts. The motility map can then be generated by invoking a sub-Riemannian constraint on the aggregate system motion under which the position velocity induced by a given shape velocity is that which minimizes the power dissipated via friction.
The locomotion of such systems is ``geometric'' in the sense that the final position reached by the system  depends only on the sequence of shapes that the system passes through, but not on the rate with which the shape changes are made.

In this paper, we consider a far more general class of systems in which the drag may be not only \emph{anisotropic} (with different coefficients for forward/backward and left/right motions), but also \emph{asymmetric} (with different coefficients for forward and backward motions).
Formally, including asymmetry in the friction replaces the Riemannian metrics on the body parts with Finsler metrics. We demonstrate that the sub-Riemannian approach to constructing the system motility map extends naturally to a sub-Finslerian approach and identify system properties analogous to the constraint curvature of sub-Riemannian systems that allow for the characterization of the system motion capabilities.

\end{abstract}

\maketitle
\section{Introduction}

Geometric treatments of locomotion provide insight into how robots and animals convert internal shape changes into displacement through the world.
These insights include:
\begin{enumerate}
\item Understanding the nature of optimal gait cycles for systems with different body types and in different environments, along with commonalities of optimal gaits across these different systems

\item Identifying low-dimensional structures within the dynamics of systems moving in complex media (e.g., sand), which simplify the analysis of such systems.
\end{enumerate}

These geometric frameworks are built on a foundation of differential geometry and Lie group theory.
In the traditional geometric locomotion approach, we consider the motion of a system whose interactions with the environment
are encoded by a \emph{Riemannian metric} describing the rate at which energy would be dissipated if the system moves with some shape, shape-change rate (``shape velocity''), and body velocity.
Combining this metric with a least-constraint principle then generates a constraint %
on the system motion, under which commanded shape velocities induce body velocities that minimize the work done by dissipation forces.
The resulting relationship is often called the ``local connection'' or the ``motility map''. %

In geometric gait analysis, the net displacement over a gait (cyclic change in shape) is determined by the \emph{curvature} of these %
power-minimizing constraints, which can be expressed in terms of derivatives of the motility map.
This curvature can be used both to identify good ``central points'' for  shape oscillations that maximize motion and to find the amplitudes and phase-coordinations that most efficiently turn these oscillations into net displacements.

\begin{figure*}
\centering
\includegraphics[width=0.95\textwidth]{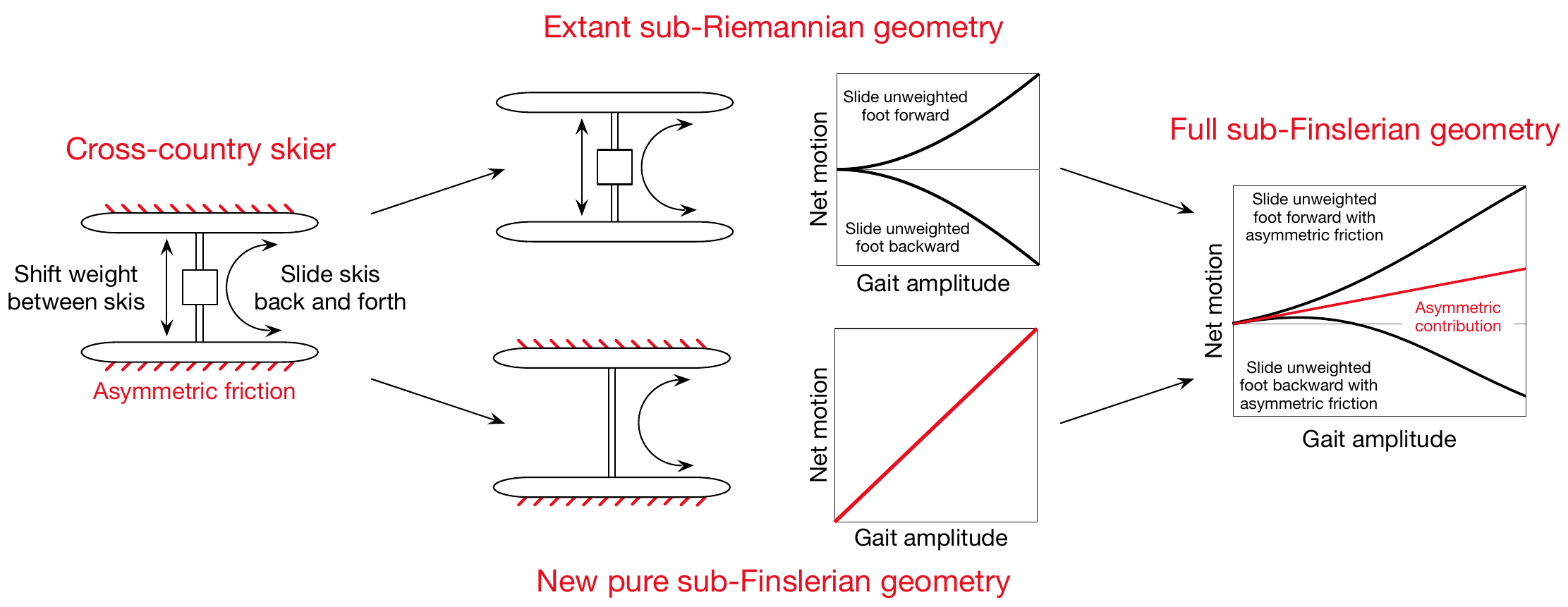}
\caption{Cross-country skier as a model system for sub-Finsler locomotion. A simple model for cross-country skiing is that the skis have asymmetric viscous friction with the ground (making it harder to slide them backwards than to slide them forwards), and that the skier can modulate this friction by shifting weight onto one or other of the skis.
\\ \vspace{1ex}
A similar skier with symmetric friction on the skis can be modeled using extant sub-Riemannian geometric methods. A key result from this prior work is that the net motion is proportional to the product of the weight and ski oscillations---growing quadratically if the oscillation amplitudes are increased together---and that its sign depends on the relative phase of the oscillations.
\\ \vspace{1ex}
Conversely, geometrically modeling a skier with only asymmetric friction requires the new sub-Finslerian analysis introduced in this paper. A key result of this analysis is that the net motion grows linearly with the sum of oscillation amplitudes, and its direction is independent of the relative phases of the oscillations.
\\ \vspace{1ex}
Combining these effects produces a locomotion model in which the full skier can use weight-shifting to boost or counteract the positive motion induced by the asymmetric friction.}
\label{fig:IntroSkier}
\end{figure*}

A limitation of the standard geometric locomotion framework is that it relies on the existence of the Riemannian metric that determines system dynamics.
Such metrics exist for systems that experience linear dissipative forces at the contact points between the system and its environment.%
\footnote{A geometric locomotion model can also be constructed for systems whose dynamics are determined by conservation of momentum (for which the Riemannian metric is the system inertia tensor), but in this paper we restrict our attention to systems with dissipative environmental forces.}
A Riemannian metric can also provide a good approximation for dynamic Coulomb friction.
The forces may be anisotropic (meaning that the drag coefficients in some directions of contact motion may be greater than in others), but because they are linear, they must be both positively homogeneous and symmetric---flipping the sign of the sliding velocity of a contact  flips the sign of the resulting drag force but leaves its magnitude unchanged.

These conditions appear directly in the equations of motion for systems moving in high-viscosity fluids, and have been demonstrated to be a useful approximation for systems moving in sand or other granular media~\citep{Hatton:2013PRL,doi:10.1073/pnas.2320517121}.

However, they are poor models for systems with directional gripping elements such as claws or scales that can glide easily in one direction, but ``catch''  strongly when the direction of relative motion is reversed.
Complex foot structures with highly asymmetric friction are widespread across terrestrial organisms at all scales regardless of evolutionary clade.
Elaborate claws and directional adhesives appear both in insects~\citep{gorb2001attachment} and in geckos~\citep{naylor2019attachment}.
Many types of locomotor claws appear in
arachnids~\citep{shultz1989morphology}.
Larger and more familiar claws contribute to the locomotion and attachment of  birds~\citep{tinius2017points}, lizards~\citep{zani2000comparative, crandell2014lizardclaws}, and mammals~\citep{hamrick2001development, manning2006mammalclaws}.
Directional claws even appear in more exotic clades such as tardigrades~\citep{suzuki2022beautiful} and velvet worms~\citep{blaxter2011velvet}.
This ubiquity of asymmetric contact friction in the animal world provides strong evidence that foot structures with directional properties provide a significant benefit to terrestrial locomotion in general, since it likely represents convergent evolution exploiting some underlying functional advantages.
We propose avoiding the complexities of modeling the states of the ``small'' DoF such as individual insect claws or individual gecko setae by a lumped-parameter approach -- modeling the entire contact as having directionally dependent friction.

In this paper, we propose an extension and generalization of the geometric locomotion framework to account for more general contact forces, including the asymmetric case.
The key elements of this generalization are:
\begin{enumerate}
\item We replace the Riemannian metrics in the dynamics formulation with Finsler metrics, which admit asymmetric costs while preserving many other properties of Riemannian metrics.
\item We identify a means of constructing sub-Finslerian locomotion constraints by combining the Finsler metrics with the least-constraint principle used previously.
\item We develop a notion of curvature for sub-Finslerian constraints that augments the standard sub-Riemannian curvature with ``ratcheting'' terms that capture the net effect of oscillating a system's shape in the presence of asymmetric friction.
\end{enumerate}

We present this proposed generalization via a set of tutorial examples, first reviewing the fundamental principles of sub-Riemannian locomotion, and then examining how the associated mathematical objects change when the Finsler metric is introduced. %
For these examples, we have constructed a minimal working model inspired by the mechanics of cross-country skiing.
Our model skier, illustrated in \zcref[S]{fig:IntroSkier}, consists of two skis that can slide forward and backward relative to each other and a weight that can be shifted between the skis. The skis are subject to viscous friction with the ground, and the coefficient of this friction scales with the proportion of the weight that each ski is supporting. Additionally, the viscous coefficient for sliding the skis backwards may be greater than for sliding the skis forwards.

Reducing this system into simpler systems that rely only on weight-shifting or asymmetric friction for forward propulsion allows us to identify locomotion principles that
\begin{enumerate}
\item Carry over directly from classical sub-Riemannian geometry (e.g., that part of the net motion induced by a gait cycle is quadratic in the amplitude of the gait and has a sign that depends on the order in which the steps of the cycle are completed).
\item Are specifically new to the sub-Finslerian analysis (e.g., that the magnitude of the asymmetric contribution to net motion is linear in the amplitude of the gait, and independent of the cyclic order).
\end{enumerate}
Recombining these principles produces a model for the locomotion of the full skier, in which the linear contribution from asymmetric friction dominates the net motion produced by small-amplitude gaits, and the quadratic contribution from weight-shifting dominates at large amplitudes.

\subsection*{Related work}

Our system treatment here builds on a Riemannian geometric treatment of locomotion under anisotropic (but symmetric) friction, developed over the past four decades by various groups (including our own) in the physics, mathematics, and robotics communities.
The core insight treating locomotion using Gauge theory appeared in \citet{Shapere:1989a}, with \citet{Murray:1993, walsh95} expressing these ideas in a finite dimensional form and showing the importance of Lie brackets in capturing key effects, and \citet{ostrowski98a} focusing on undulatory locomotion.
The works of \citet{Melli:2006, Morgansen:2007}, and \citet{Shammas:2007} focused on planning and control.
Optimal motion was the focus of \citet{Avron:2008} and \citet{Alouges:2008}.
A key innovation came from \citet{Hatton:2011IJRR, Hatton:2015EPJ} which showed the importance of coordinate choice when applying geometric locomotion theory to finite-amplitude system motions.
Modern treatments of gait optimization using motility maps can be found in \citet{DeSimone:2012aa} and \citet{Ramasamy:2019aa}.
Data driven approaches to this optimization can be found in \citet{Bittner2018ab}, with extensions for low but non-zero Reynolds numbers \citep{Kvalheim2019gmpsr}, underactuated systems \citep{bittner-2020-SUDS-BioB}, and more modern machine-learning function fitting frameworks \citep{hu2025learning}.

Finsler metrics have received some attention in robotics, e.g.,~\citet{Bidabadi:2010aa,Ratliff:2021aa,Monforte:02,Van-Wyk:2022aa}, but to the best of our understanding this attention has been directed at path-planning problems where the cost of following a path depends on the direction in which it is traversed, and not at constructing locomotion models induced by Finslerian dynamics. The mathematical literature contains numerous works on Finsler geometry, including connection forms on Finsler manifolds, but our search of the literature has not identified any works considering the local forms of such connections.

Conversely, the notion of exploiting asymmetric (as opposed to merely anisotropic) friction has appeared repeatedly in the robotics and biological literature, e.g., \citet{Menciassi:2006aa,Branyan:2020wb,das2023earthworm,Wong:2023aa,Tirado:2025aa}
but such friction does not appear to have been linked to the notion of a Finsler metric.

\section{A Review of Classical Geometric Mechanics for Locomotion}
\label{sec:geomechoverview}

The core idea in applying geometric mechanics to mobile systems is that a mobile robot or animal can be modeled as a collection of $m$ rigid bodies.
Intuitively, the overall location and orientation of the ensemble in the world is its \emph{position}, and the relative location and orientation of the constituent bodies is the system's \emph{shape}.\footnote{Mathematically speaking, position and shape arise from an underlying symmetry represented by a group that acts on configurations.
Our intuitive notion of shape is that which arises from the group of rigid body motions, which by its group action defines that any two configurations that are related by a rigid body motion represent the same shape.
However, a notion of shape can arise from other (Lie) groups just as well.}

The interactions between the system and its environment depend on the motion of each link as seen in its own local directions, possibly influenced by the relative positions and velocities of the other links.
The position changes induced by changing shape can be found by pulling back the environmental interactions into the position-and-shape configuration space, and then solving an equilibrium or least-constraint problem.

In our present review of geometric mechanics principles, we provide a general form of the key equations and specific instantiations of the equations for a planarized version of the skier from \zcref[S]{fig:IntroSkier}.
This choice of example allows us to include full expressions for the dynamics and show the essential aspects of sub-Riemannian and sub-Finslerian locomotion, while eliding some details arising from the non-commuting interaction of translations and rotations which we have explored in previous publications~\citep{Hatton:2015EPJ,Ramasamy:2019aa, Bass:2022wn}. %

\subsection{Kinematics}
A convenient means of representing the system kinematics is to separate the coordinates so that we identify a \emph{body frame} for the system (e.g., a base link or a center-of-mass frame) whose position  $\fiber\in SE(n)$, $n=2 \text{~or~} 3$, serves as the position of this frame as the system frame, and a set of parameters $\jointangle$ (e.g., the joint angles between adjacent links) to describe the shape of the system, which do not change if the body is moved without a change in shape.
These shape parameters can then be used to describe the position of each link relative to the body frame as $\altfiber_{i}(\alpha)\in SE(n)$, such that the position of the link in the world, $\fiber_{i}$, is the composition of the system position with this relative position, $\fiber_{i}(\fiber, \jointangle)= \fiber\,\altfiber_{i}(\jointangle)$.\footnote{See the Appendix of~\cite{Ramasamy:2019aa} for details of these kinematics, and for an analogous construction for continuum systems.}

\begin{figure}
\centering
\includegraphics[width=0.45\textwidth]{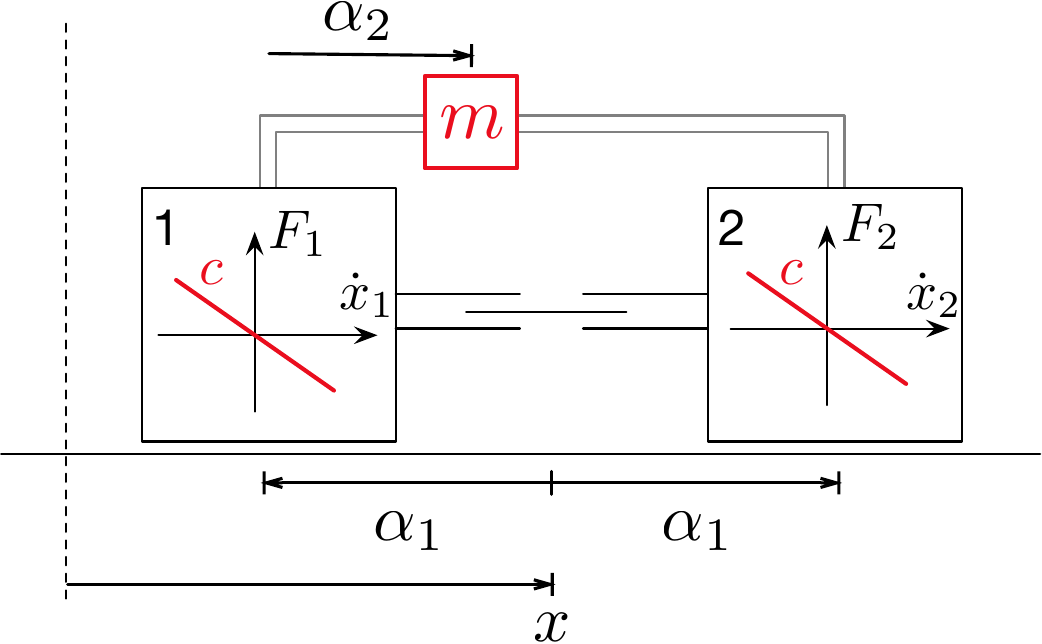}
\caption{Planarized cross-country skier model. The two blocks in contact with the ground are each able to slide along the $x$ axis and are connected by a linear actuator. A second block (which contains all of the system mass) is connected to a rail between these blocks. The two blocks are subject to viscous friction, whose coefficient depends both on a base value $\mu$ and the proportion of the sliding mass whose weight the friction-block is supporting.}
\label{fig:RiemannSkier}
\end{figure}

\begin{example}

The system in \zcref[S]{fig:RiemannSkier} is a planarized version of the cross-country skier from \zcref[S]{fig:IntroSkier}, consisting of two blocks connected by a linear actuator and a weight sliding on a rail between them. Its position $\fiber=(x)$ is the mean of the two block positions, and its shape $\jointangle$ is the $\jointangle_{1}$ half-distance between the blocks and the $\jointangle_{2}$ placement of the sliding mass, such that the positions of the two blocks in terms of generalized coordinates are
\begin{align}
\fiber_{1} &= (x_{1}) = (x - \jointangle_{1})
\\
\intertext{and}
\fiber_{2} &= (x_{2}) = (x + \jointangle_{1}),
\end{align}
with $\jointangle_{2}$ taking on values of $0$ and $1$ when positioned over the rear or forward block, respectively.
Note that the body position is intentionally not the position of the center of mass.
Although using the CoM frame as position is often convenient, it is not required.
\end{example}

By extension, movement of the system bodies can be described by the system's \emph{body velocity}, $\bodyvel$---the velocity of the body frame with respect to the fixed reference frame, but expressed in local coordinates aligned with the body frame's position and orientation---and its \emph{shape velocity}, $\jointangledot$---the rate at which the shape parameters are changing.
Differentiating the system kinematics\footnote{\emph{Id}, see the Appendix of~\cite{Ramasamy:2019aa} for details.} allows us to construct a \emph{body Jacobian} for each of the links, mapping the system body velocity and shape velocity to the body velocities of the individual links (which are expressed in coordinates rigidly attached to the links),
\begin{equation}
\bodyvel_{i} = \jac^{b}_{i}(\jointangle) \begin{bmatrix}
\bodyvel \\ \jointangledot
\end{bmatrix}.
\end{equation}

\begin{example}
The body velocities of the two blocks are the time derivatives of their positions in the world, and the Jacobians from generalized-coordinate velocities to block velocities can be found by taking the derivatives of the block positions with respect to corresponding coordinates,
\begin{align}
\fibercirc_{1} &= \begin{bmatrix} \xdot_{1} \end{bmatrix} = \overbrace{\begin{bmatrix} 1 & -1 & \pmi0 \end{bmatrix}}^{\jac^{b}_{1}} \begin{bmatrix} \xdot \\ \jointangledot_{1} \\ \jointangledot_{2} \end{bmatrix} \label{eq:block1Jacobian}
\\
\intertext{and}
\fibercirc_{2} &= \begin{bmatrix} \xdot_{2} \end{bmatrix} = \overbrace{\begin{bmatrix} 1 & \pmi1 & \pmi0 \end{bmatrix}}^{\jac^{b}_{2}} \begin{bmatrix} \xdot \\ \jointangledot_{1} \\ \jointangledot_{2} \end{bmatrix}.\label{eq:block2Jacobian}
\end{align}
\end{example}

A key feature of this construction for formal treatments is that expressing both the system and link velocities in their own local directions makes the body Jacobians independent of the system position $\fiber$ even when mixed translations and rotations are present.\footnote{This property is formally part of the \emph{left-invariance} of the system kinematics. Continuing in this formal vein, the body velocities $\bodyvel$ and $\bodyvel_{i}$ can all be treated as elements of the Lie algebra of $SE(n)$, and the columns of $\jac^{b}_{i}(\jointangle)$ are generated from from the adjoint actions of the transforms between the body or joint frames and the $i$th link frame, as detailed in the appendices of~\citet{Ramasamy:2019aa}.}

\subsection{Local physics}
A common model for the physics of locomotion is that the motion of each link is resisted by the surrounding environment. This resistance may be greater in some local directions than in others (e.g., an elongated shape sliding more easily along its long axis as compared to crosswise), and may also depend on the system shape (e.g., weight distribution may affect the normal force at a link, the presence of a nearby link may change the fluid flow around it, or a three-dimensional system in contact with a two-dimensional surface may change the angle-of-attack between its contact bodies and the surface).

Within this general class of resistive models, \emph{linear-viscous} models play a particularly prominent role in the locomotion literature: they arise naturally from the study of micro-organisms, have a mathematical structure that is inherently differential-geometric, and provide an approximate model for systems whose ``true" physics is not conducive to analysis~\citep{Hatton:2013PRL,doi:10.1073/pnas.2320517121}. In these models, the resistive force acting on a link (as expressed in link-aligned coordinates) is found by multiplying a matrix of drag coefficients into the link's body velocity,
\begin{equation} \label{eq:linkforce}
\bodyforce_{i} = -\dragmetric_{i}(\jointangle)\, \bodyvel_{i}
\end{equation}
(where the ``undercircle" in $\bodyforce_{i}$ indicates that it is in the same frame as the body velocity $\bodyvel_{i}$). This resistive force is not necessarily aligned with the link's direction of motion; in particular, if a link has ``large" and ``small" drag directions and moves in a direction slightly off from the ``small" drag direction, it will experience a ``deflecting force" aligned towards the large-drag direction.

An alternative expression of linear-viscous physics is that the power dissipated when moving a link is quadratic in the link velocity, and can be calculated as the drag-modulated inner product of the link's body velocity with itself,
\begin{equation} \label{eq:linkpower}
P_{i} = \transpose{\bodyvel_{i}} \dragmetric_{i}(\jointangle)\, \bodyvel_{i}.
\end{equation}
This expression highlights the differential-geometric nature of the viscous drag matrix $\dragmetric$: It serves as a Riemannian metric on the link's configuration space, which can either dualize a velocity into its corresponding force, as in~\eqref{eq:linkforce}, or provide an energy-based norm on the link velocity that accommodates mixed-coordinate velocities (e.g., translation and rotation components of $\bodyvel_{i}$) as in~\eqref{eq:linkpower}.

\newcommand{\blockonescale}{(1-\jointangle_{2})^{2}}
\newcommand{\blocktwoscale}{(\jointangle_{2})^{2}}
\begin{example}
In our skier model the skis are subject to viscous drag that is linearly proportional to their velocity and depends quadratically on the proportion of the skier's weight that is supported by that ski.\footnote{We have selected this quadratic function because it provides just enough complexity to allow us to highlight some nonlinear properties of the geometric locomotion model in later examples. It is also physically motivated by the mechanics of cross-country skiing: traditional skis are ``cambered" into an arch, and their undersides have more grip at the center than at the ends. Placing weight onto a ski flexes it, increasing the contact between the high-grip center section and the snow. The ski is geometrically stiffer with respect to load when it is arched than when it is flat (i.e., the ski is a ``softening spring"), so the increase in grip grows superlinearly with the proportion of the weight centered on the foot, and our quadratic model is a minimal example of this class of functions.} In this model the (1x1) drag matrices on the blocks are:
\begin{align}
F_{1} &= -\overbrace{m\fricco[ \blockonescale ]}^{\dragmetric_{1}}\, \xdot_{1} \label{eq:block1Riemanmetric}
\\
\intertext{and}
F_{2} &= \smash{-\overbrace{m\fricco[ \blocktwoscale]}^{\dragmetric_{2}}}\, \xdot_{2}, \label{eq:block2Riemanmetric}
\end{align}
and the power required to move each block can be readily computed by multiplying its drag matrix by the square of the block's velocity.
\end{example}

\subsection{System physics}
The drag metrics on the individual link motions can be transformed (pulled back) via the link Jacobians into drag metrics on the system motions,
\begin{equation}\label{eq:Riemannpullback}
\dragmetric(\jointangle) = \sum_{i} \jac^{b,T}_{i}\, \dragmetric_{i}\, \jac^{b}_{i},
\end{equation}
which, as in the local case, can be used both to dualize a system velocity vector to the resulting resistance forces acting on the body frame and the shape variables (e.g., joint torques),
\begin{equation} \label{eq:systemforce}
\begin{bmatrix} \bodyforce \\ \worldforce_{\jointangle}\end{bmatrix}
=
-\overset{\dragmetric(\jointangle)}{\begin{bmatrix}
\dragmetric_{\fiber\fiber} & \dragmetric_{\fiber\jointangle} \\
\dragmetric_{\jointangle\fiber} & \dragmetric_{\jointangle\jointangle}
\end{bmatrix}}
\begin{bmatrix}
\bodyvel \\ \jointangledot
\end{bmatrix}
\end{equation}
and to find the power dissipated when moving the system,
\begin{equation} \label{eq:systempower}
P = \begin{bmatrix}
\transpose{\bodyvel} & \transpose{\jointangledot}
\end{bmatrix}
\dragmetric(\jointangle)
\begin{bmatrix}
\bodyvel \\ \jointangledot
\end{bmatrix}.
\end{equation}

\newcommand{\blocksumnegative}{2\jointangle_{2}-1}
\newcommand{\blocksumpositive}{2\jointangle_{2}^{2}-2\jointangle_{2}+1}
\begin{example}
The system metric for the planarized skier can be found by pulling back the local metrics from \zcref[S]{eq:block1Riemanmetric,eq:block2Riemanmetric} through the Jacobians from \zcref[S]{eq:block1Jacobian,eq:block2Jacobian},
\begin{align}
\begin{split}
\dragmetric &= \begin{bmatrix} \pmi1 \\ -1 \\ \pmi0 \end{bmatrix}
mc[\blockonescale]
\begin{bmatrix} 1 & -1 & \pmi0 \end{bmatrix}\\
&\qquad +
\begin{bmatrix} 1 \\ 1 \\ 0 \end{bmatrix}
mc[\blocktwoscale]
\begin{bmatrix} 1 & 1 & 0 \end{bmatrix}
\end{split}
\\[2ex]
&= m c \begin{bmatrix} \blocksumpositive & \blocksumnegative & 0 \\ \blocksumnegative & \blocksumpositive & 0 \\
0 & 0 & 0\end{bmatrix}.\label{eq:skierdragmetricRiemann}
\end{align}
The first two diagonal terms of the system metric are always positive, indicating that system position and inter-ski velocity are always met with direct resistance from the block friction; the last diagonal value being zero corresponds to the system being able to freely move the weight between skis. The $\jointangle_{2}$-dependence of the off-diagonal terms means that when the mass is centered between the blocks (sending the term to zero), the position and shape are decoupled, e.g., such that bulk position motion exerts no force on the shape mode, but that away from this position, the position and shape modes are coupled, e.g., because having mass on the back block makes it drag more, so that drag from system forward velocity will pull the blocks apart.
\end{example}

\subsection{Power-minimizing constraints} \label{sec:subriemannconds}
For systems operating in high-drag environments, two conditions can be used to construct a locomotion model: First, that the system is driven only through forces applied at the joints, with no $\bodyforce$ ``direct thrust" applied to the body frame, and second, that
the viscosity-to-mass ratio of the system is large relative to the frequency with which the shape variables change their motion, such that the
system can be considered as always moving at terminal velocity, with no second-order inertial effects.

Taken together, these conditions mean that the system velocities are constrained to those for which the $\bodyforce_{i}$ link forces are in equilibrium around the body frame. This condition can be encoded by extracting the net force on the body frame from the top row of~\eqref{eq:systemforce} and setting it equal to zero,
\begin{equation}
\bodyforce=-\begin{bmatrix} \dragmetric_{\fiber\fiber} & \dragmetric_{\fiber\jointangle} \end{bmatrix} \begin{bmatrix}
\bodyvel \\ \jointangledot
\end{bmatrix} =\mathbf{0}.
\end{equation}
Formally, the top row of $\dragmetric$ serves as a \emph{Pfaffian constraint} (a linear map whose nullspace contains velocities compatible with the system constraints), here encoding the restriction to force-equilibrium trajectories.

Rearranging the components of the constraint matrix then allows us to to construct configuration velocities that satisfy the corresponding condition by selecting an $\jointangledot$ shape velocity component and finding its complementary position velocity in the null space of the top block of $\dragmetric$,
\begin{equation} \label{eq:motility}
\bodyvel = \overbrace{(-\inv{\dragmetric_{\fiber\fiber}}\dragmetric_{\fiber\jointangle})}^{\mixedconn(\jointangle)}\,\jointangledot.
\end{equation}
The \emph{motility map} $\mixedconn(\jointangle)$ is analogous to the Jacobian of a robot arm, mapping directly controlled joint velocities into indirectly controlled rigid-body motion.%
\footnote{In much of the geometric mechanics literature, what we refer to as the ``motility map'' is often called the ``(negative of the) local connection''.
This terminology reflects its mathematical provenance from the fiber bundle literature, and not its application.
We have therefore chosen a more readable name for use here.}

An alternative construction of the motility map is based on a least-constraint principle, in which the $\bodyvel$ associated with a given $\jointangledot$ velocity is the $\bodyvel$ that minimizes the dissipated power $P$ out of all configuration velocities that contain $\jointangledot$.
In the linear case considered above, the two constructions are directly equivalent---the derivative of $P$ in~\eqref{eq:systempower} with respect to $\bodyvel$ is
\begin{equation}
\frac{\partial P}{\partial \bodyvel} = \begin{bmatrix} \dragmetric_{\fiber\fiber} & \dragmetric_{\fiber\jointangle} \end{bmatrix} \begin{bmatrix}
\bodyvel \\ \jointangledot
\end{bmatrix},
\end{equation}
which differs only by a sign from the body force equation, and is thus zero when $\jointangledot$ and $\bodyvel$ are related as in~\eqref{eq:motility}.

\begin{figure}
\centering
\includegraphics[width=.9\linewidth]{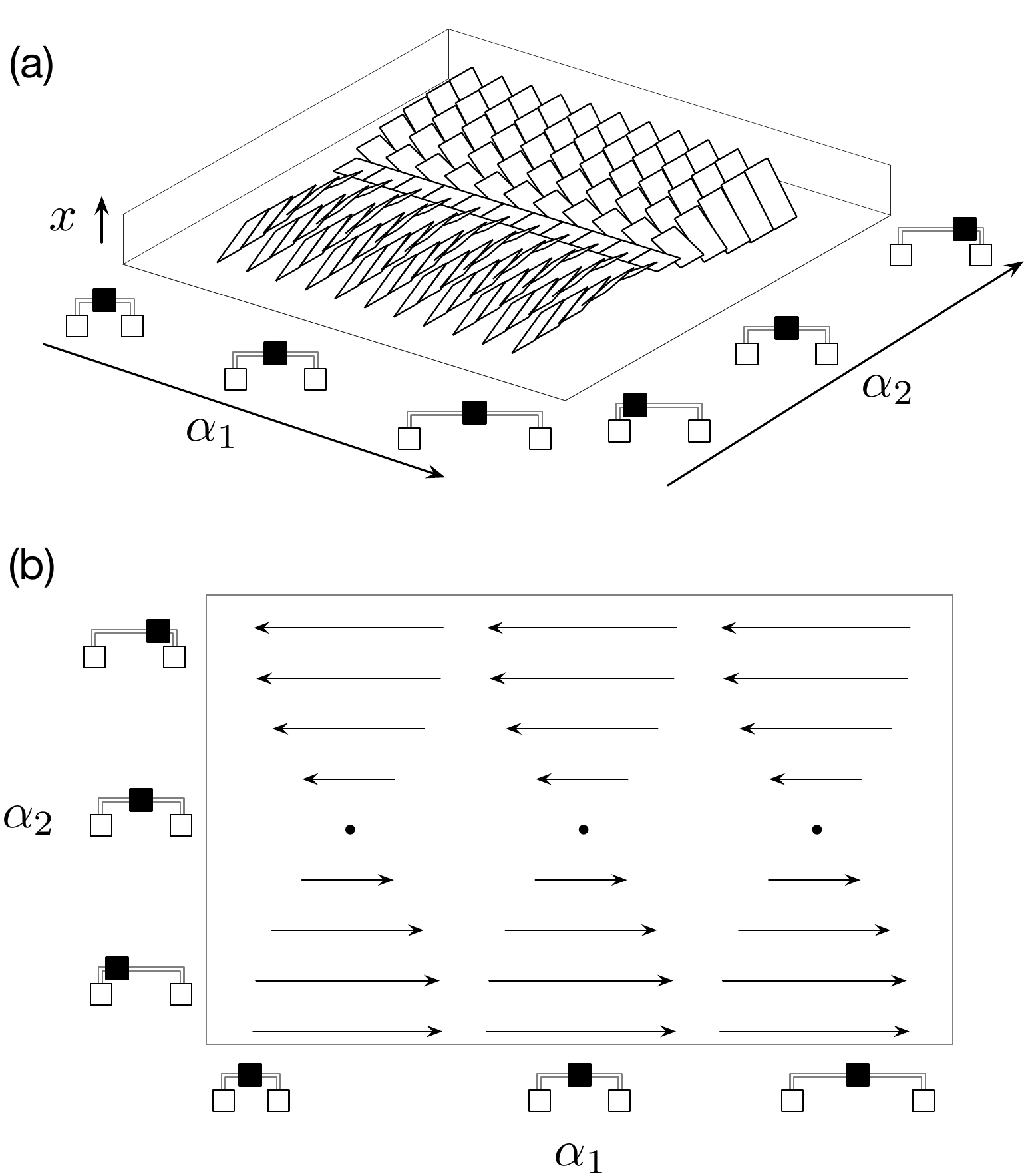}
\caption{The motility map $\mixedconn$ for the planarized skier encodes a constraint that restricts the system velocity to a two-dimensional subspace of the possible $(\dot{x},\dot{\alpha})$ values in the tangent space at each point, as described in \zcref[S]{exp:planarizedskiermotilitymap}. The fundamental geometry of this constraint is captured by the plane field in (a), in which each plane contains all the allowable directions of configuration velocity at each point. It is often more convenient to instead visualize the constraint as an arrow field as in (b), in which the arrows point in the ``uphill'' direction of their corresponding planes, with length proportional to the planes' steepness.}
\label{fig:constraintfield}
\end{figure}

\begin{example} \label{exp:planarizedskiermotilitymap}
The motility map for the planarized skier can be found by multiplying the negative inverse of the top-left entry of $\dragmetric$ from \zcref[S]{eq:skierdragmetricRiemann} into the rest of the top row of the metric matrix. The $mc$ coefficients are self-canceling in this multiplication, leaving the motility map as
\begin{equation} \label{eq:planarizedskiermotility}
\xdot = \overbrace{\begin{bmatrix} -\dfrac{\blocksumnegative}{\blocksumpositive} & 0 \end{bmatrix}}^{\mixedconn(\jointangle)} \begin{bmatrix} \jointangledot_{1} \\ \jointangledot_{2} \end{bmatrix}.
\end{equation}
This motility map has no $\jointangle_{2}$ component and a $\jointangle_{1}$ component that is positive for $\jointangle_{2}<\frac{1}{2}$ (because the back block has more grip, so expansion pushes the system forward) and negative for  $\jointangle_{2}>\frac{1}{2}$ (because the front block has more grip, so expansion pushes the system backwards).

This motility map is illustrated in \zcref[S]{fig:riemanngeomech}(a) as a plane field, in which each plane contains the set of vectors satisfying the
\begin{equation}
\mathbf{0} = \bodyforce =  \tfrac{\partial P}{\partial \bodyvel} =\begin{bmatrix} \blocksumpositive & \blocksumnegative & 0 \end{bmatrix} \begin{bmatrix} \xdot \\ \jointangledot_{1} \\ \jointangledot_{2} \end{bmatrix}
\end{equation}
nullspace condition on the system's physically allowable motions. When the $\jointangledot$ shape velocity is taken as an independent input, the planes act as ``ramps'' dictating the dependent $\xdot$ position velocity that keeps the system in the force-equilibrium null space. These fields are sloped ``up'' with respect to $\jointangle_{1}$ when $\jointangle_{2}$ is positive (and vice versa), corresponding to how the position of the weight affects the coupling between extension/contraction and the movement of the center of the system.

Because accurately drawing and reading plane fields is difficult, it is often more convenient to illustrate the motility map via arrow fields as in \zcref[S]{fig:riemanngeomech}(b). In this plot, each arrow points ``up'' the slope of the corresponding plane, with length proportional to its steepness. Equivalently, the components of the arrows in the $\jointangle_{1}$ and $\jointangle_{2}$ directions are the steepness of the planes along those two directions, with component length equal to the corresponding components of $\mixedconn$. The $\xdot$ position velocity induced by moving with a given $\jointangledot$ shape velocity (and thus satisfying the nullspace constraint) is then the ``dot product'' of the slope vector with the shape velocity, corresponding to the computation in \label{eq:planarizedskiermotility}. The arrow plot in particular highlights the property that the steepness of the planes changes rapidly in the middle of the $\alpha_{2}$ range%
\end{example}

\subsection{Gaits} \label{sec:gaits}
Unlike vehicles, which can generally spin their wheels without limits, locomoting systems such as snakes and limbed crawlers are typically subject to joint limits, and must move their joints within these limits.
Planning and control for such systems typically involves identifying \emph{(periodic) gaits}---cyclic shape changes---that produce net displacement in a set of different directions, which can then be chained together into a motion sequence.

\begin{figure}
\centering
\includegraphics[width=0.45\textwidth]{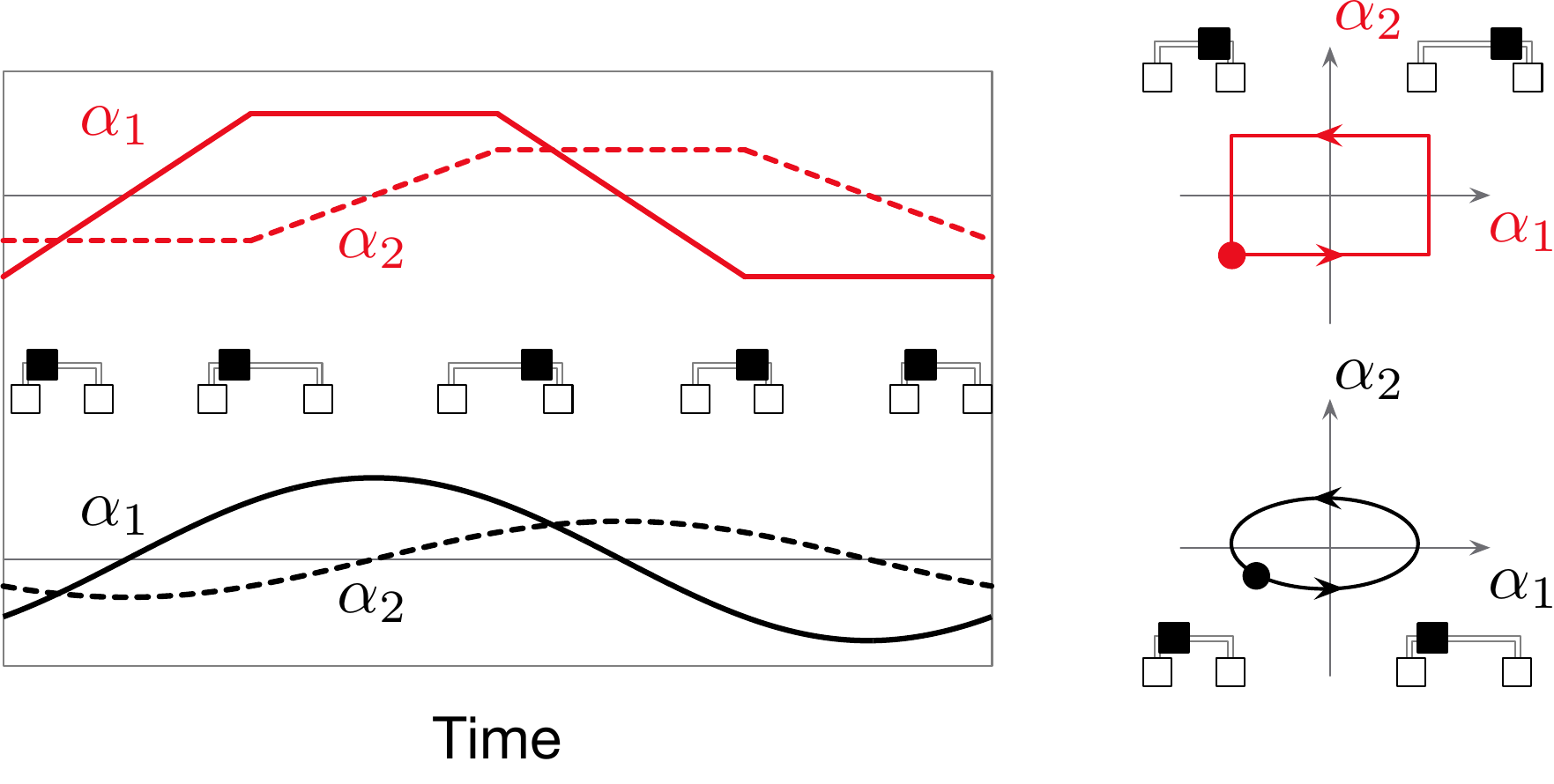}
\caption{Gaits are defined by the amplitudes, relative phases, and shapes with which their degrees of freedom oscillate, as described in \zcref[S]{exp:gaitgeometry}. (top) Individually moving the degrees of freedom produces a ``box'' gait. (bottom) Simultaneously moving the degrees of freedom in sinusoidal patterns produces an ``elliptical'' gait.}
\label{fig:gaitgeometry}
\end{figure}

System gaits can be described by the waveforms, amplitudes, and relative phasing of the individual joint oscillations. For geometric analysis, however, it is often more useful to think of them in terms of the trajectory that they trace out within the shape space

\begin{example} \label{exp:gaitgeometry}
Two classes of gaits that commonly appear as benchmarks when studying locomotion are ``trapezoidal wave'' gaits, in which the one joint at a time is ``unlocked'' and moved from one extreme of its stroke to another, and ``sinusoidal wave'' gaits, in which the degrees of freedom are oscillated sinusoidally. As illustrated in \zcref[S]{fig:gaitgeometry}, these gaits respectively trace out boxes and ellipses in the shape space.
\end{example}

The displacement induced by executing a gait is determined by how the system ``climbs'' or ``descends'' the constraints between shape and position velocities as it changes shape.

\begin{figure}
\centering
\includegraphics[width=0.45\textwidth]{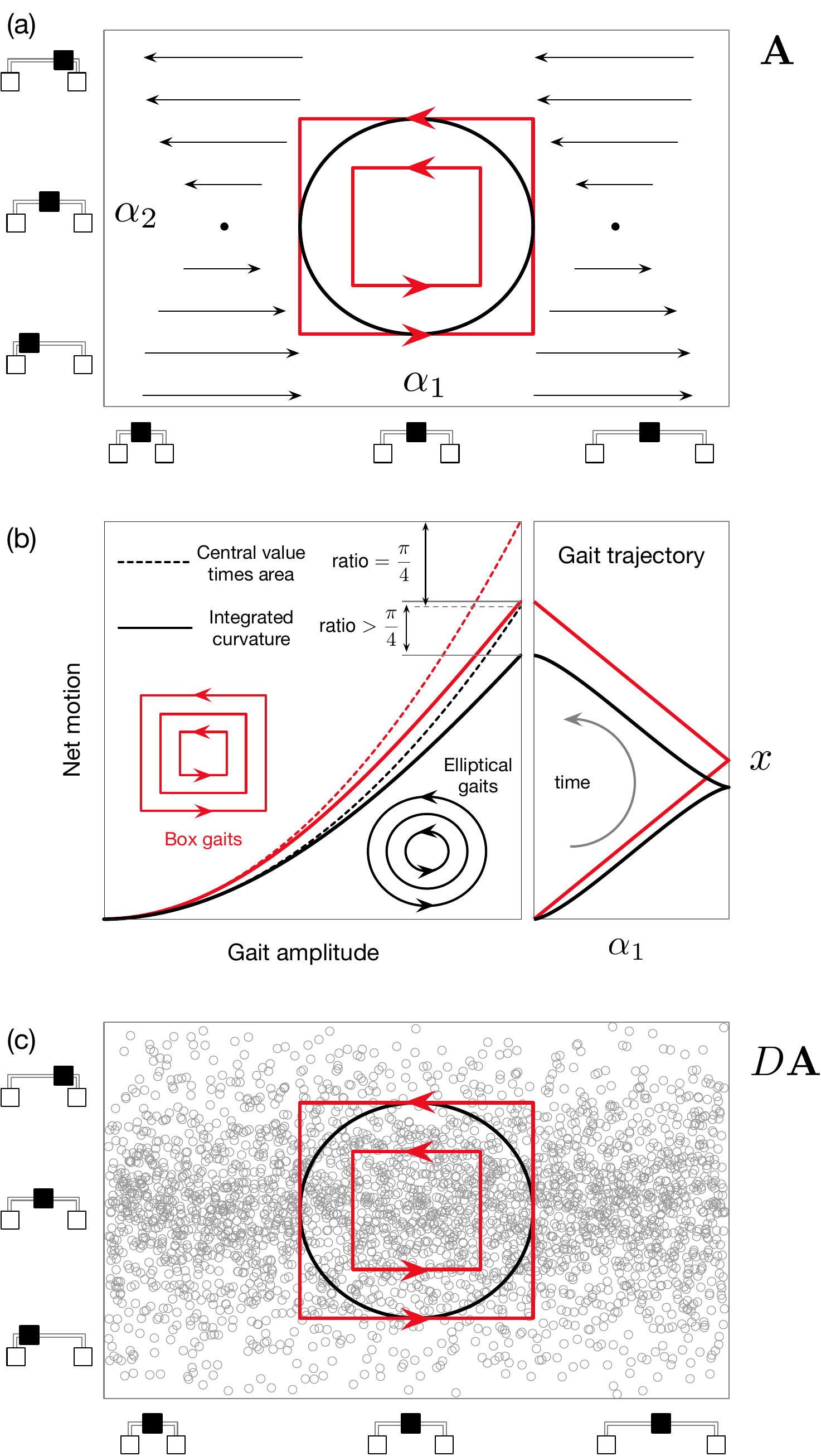}
\caption{Locomotion of the planarized skier. (a) The skier's motility map $\mixedconn$ evaluated over its shape space, together with a pair of box-shaped gaits at different scales and an elliptical gait with the same amplitude as the bigger box. The small cartoons of the skier along the axes illustrate the geometric meaning of the $\jointangle_{1}$ and $\jointangle_{2}$ variables. (b) The net displacement induced by the gaits grows approximately quadratically with the gait amplitude, with an approximately $\pi/4$ ratio between the box and elliptical gaits. (c) The constraint curvature $D\mixedconn$ (plotted with density of the stippling points indicating its magnitude). The slight subquadratic growth in the net displacement corresponds to bigger gaits enclosing regions where $D\mixedconn$ is less rich, and the $>\pi/4$ ratio between displacement from large ellipses and large boxes corresponds to the areas ``given up" by the ellipses being biased towards the low-richness regions of the shape space.}
\label{fig:riemanngeomech}
\end{figure}

\begin{example}
In the ``box" gaits illustrated in \zcref[S]{fig:riemanngeomech}(a), the skier first puts its weight on the back block and expands, which causes the $x$ position of the system center to move forward in proportion to the increase in $\jointangle_{1}$, as illustrated in the right panel of \zcref[S]{fig:riemanngeomech}(b). It then shifts the weight to the front block, so that contracting $\jointangle_{1}$ to the original size continues to move the system forward (rather than sending it back to its original position, as would have happened if the weight had remained on the rear block).

The elliptical gaits function similarly to the box gaits, except that the block spacing and weight position are oscillated sinusoidally instead of ensuring that the weight is at its extremal position before expanding or contracting the joint. This change in synchronization means that some of the $\jointangle_{1}$ motion happens when it is more weakly coupled with $x$, producing a ``cusped" trajectory with less net position motion for the same amplitude of shape change, as illustrated at the right of \zcref[S]{fig:riemanngeomech}(b).\footnote{Note that this does not necessarily mean that the box gait is \emph{faster}: If we match the average speed of expansion and contraction for a pair of gaits and it takes time to shift the weight between blocks, the cycle-time savings from changing both shape variables simultaneously may lead to an overall faster position speed than produced by waiting for the weight to finish moving. Conversely, if we match the cycle time of the two gaits, the elliptical gait moves its joints more slowly (and thus with less power expenditure) because it spreads this motion over a longer time interval. In this paper, we focus our attention on displacement (not speed or efficiency), but see~\citet{Ramasamy:2019aa} for some of our previous discussion of this topic.}
\end{example}

The net displacement induced by a gait depends on three key factors:
\begin{enumerate}
\item How much the motility map changes across the gait (such that the displacement induced as the system moves away from its initial shape is not canceled as it returns to that shape);
\item The amplitude of the shape changes (bigger shape changes induce proportionally bigger position motions); and
\item The orientation of the cycle (reversing the direction of a gait flips its coupling with the motility map, and so reverses the direction of the induced position motion).
\end{enumerate}
Together, the first two properties mean that the displacement induced by a gait is approximately quadratic in the amplitude of the shape oscillations---scaling up a set of oscillations provides more opportunity for the motility map to change across the gait, and for the gait to take more advantage of this change.

\begin{example} \label{exp:gaitdirectintegration}

For small-amplitude gaits with the weight near the center of the system, the change in $\mixedconn$ is approximately linear in the $\jointangle_{2}$ motion of the weight, so the net displacement induced by a box gait is approximately the product of this linear term and the $\jointangle_{1}$ range of motion of the joint. This displacement grows quadratically as the weight and joint motions are scaled up together, as illustrated in the left panel of \zcref[S]{fig:riemanngeomech}(b). For large-amplitude gaits, the displacements become subquadratic with gait scale, because the motility map changes more slowly with $\jointangle_{2}$ when the weight is close to one of the blocks.

Because the elliptical gaits perform some of their $\jointangle_{1}$ shape change without having fully moved the weight towards one block, they take less advantage of the changes in the motility map, producing the (previously noted) smaller displacement at a given amplitude, but still growing (sub)quadratically with amplitude.

Reversing the directions of the gait cycles would place the weight on the front block during expansion and the rear block during contraction, causing the system to move backward instead of forward.
\end{example}

\subsection{Geometric averaging.}
The motility-change, amplitude, and orientation contributions to net displacement can be succinctly formalized using geometric averaging theory. The two key principles in geometric averaging are
\begin{enumerate}
\item Instead of working directly in terms of the $\fiber_{\gait}$ displacement induced by a gait $\gait$, it is often cleaner to work in terms of the displacement's \emph{exponential coordinates}, $\fiberexp_{\gait}$, which describe the constant $\fibercirc$ body velocity that would bring the system from the origin to $\fiber_{\gait}$ in unit time, and so describe a ``time-normalized geometric average" of the motion during the gait.\footnote{We use an open-box accent to distinguish between Lie algebra elements that describe the exponential coordinates of a displacement ($\fiberexp_{\gait} = \log(\fiber_{\gait})$) and the present velocity of a system ($\fibercirc = \fiberinv\fiberdot$). At a casual reading level, this distinction can be elided, with the subscript highlighting the distinction between the two quantities.}

For the pure-translation motions we consider in this paper, the components of the net displacement and its exponential coordinates have the same values. For mixed translation and rotation, the flow would move the system in an arc; for example, combined forwards and clockwise velocity turns the system as it moves, resulting in net motion that is forward and to the right of the starting position.

\item The exponential coordinates for the net displacement can be approximated by calculating the \emph{curvature} of the constraints encoded in the motility map, $D\mixedconn$, and then integrating this curvature over the interior of the gait as
\begin{equation} \label{eq:constraintcurvatureintegral}
\fiberexp_{\gait} \approx \iint_{\gait} \overbrace{\extd \mixedconn + \liebracket{\mixedconn_{i}}{\mixedconn_{j>i}}}^{D\mixedconn}.
\end{equation}
As discussed below, the constraint curvature is a derivative that captures the lowest-order changes in the locomotion kinematics across different shapes, such that its integral across a cycle approximates the residual motion induced by the cycle.%
\footnote{More formally, this relationship is based on a truncated Baker-Campbell-Hausdorff formula for the gait displacement, which we discuss in detail in~\cite{Bass:2022wn}.}

\end{enumerate}

The constraint curvature has two elements: the \emph{exterior derivative} (generalized curl), $\extd \mixedconn$, which describes the \emph{nonconservativity} of the motility map (the property that passing a cyclic shape change through $\mixedconn$ can produce a non-zero arithmetic average body velocity), and the \emph{local Lie brackets}, $\liebracket{\mixedconn_{i}}{\mixedconn_{j}}$, which describe the \emph{noncommutativity} of the body-frame motions (the property that equal ``forward" and ``backward" translations with intermediate rotations don't fully cancel, so that even if a system has zero arithmetic average body velocity over a cycle, it may have a non-zero geometric average):

For each pair of shape directions $i<j$, the exterior derivative measures how each component of $\mixedconn$ changes along the other direction,
\begin{equation}
(\extd \mixedconn)_{ij} = \frac{\partial \mixedconn_{j}}{\partial \jointangle_{i}} - \frac{\partial \mixedconn_{i}}{\partial \jointangle_{j}}.
\end{equation}
This term captures the changes in $\mixedconn$ that affect the net motion over a cycle: If the $i$th component of $\mixedconn$ changes along the $j$th shape coordinate direction, then a gait that oscillates the $i$ and $j$ shape elements out of phase will produce different amounts of motion during the segments of the gait aligned with $+i$ and $-i$, and so have a non-zero average velocity, with a similar effect if the $j$th component changes along the $i$th direction. The $\extd\mixedconn$ term excludes changes in a component along the its own direction in the shape space, because these changes cancel out over a gait cycle, and so do not contribute to the average velocity.

Double-integrating $\extd\mixedconn$ over the interior of the gait reconstructs the edge-to-edge changes in $\mixedconn$ across the gait and multiplies those changes by the relevant range of shape motions to produce a (time-normalized and arithmetic) average body velocity over the gait.

For a system that can both translate and rotate, forward motions and backward motions of the same magnitude do not completely cancel each other if the system rotates between executing them (i.e., translation and rotation do not commute). For example, alternating forward and backward motions with turning motions results in sideways ``parallel-parking" motion. The local Lie bracket terms are the derivatives of the world-coordinate expressions of columns of $\mixedconn$ while moving in the directions of other columns.

These derivatives provide linearized approximation of the noncommutativity and can be used as a ``correcting term" inside the integral of $\extd\mixedconn$ to make up for some of the information that is lost by working in a body-fixed frame, thereby taking the integral in \zcref[S]{eq:constraintcurvatureintegral} from a simple arithmetic average body velocity to a better approximation of the true geometric average of the system motion during the gait.\footnote{The local Lie bracket terms capture the effects of \emph{intermediate} rotations and are distinct from the exponential-coordinates property that an average velocity of turning while translating produces an arc motion.} The Lie bracket term does not appear in any of the translation-only examples we consider in this paper, but will become important in follow-on work, so we include it here in our general equation. See~\cite{Hatton:2015EPJ, Bass:2022wn} for a more detailed description of this contribution.

\begin{example}
Because our example system's position space is only displacement in the $x$ direction (with no rotation),  its constraint curvature contains only the exterior derivative of the motility map in \zcref[S]{eq:planarizedskiermotility},
\begin{subequations}
\begin{align}
\extd \mixedconn &= (d\mixedconn)_{12}
\\
&= \frac{\partial \mixedconn_{2}}{\partial \jointangle_{1}} - \frac{\partial \mixedconn_{1}}{\partial \jointangle_{2}}
\\
&= \frac{2}{\blocksumpositive}-\frac{(\blocksumnegative)(4\jointangle_{2}-2)}{(\blocksumpositive)^2}\\
&=- \frac{4\jointangle_{2}(\jointangle_{2}-1)}{(\blocksumpositive)^{2}}.
\end{align}
\end{subequations}
As illustrated in \zcref[S]{fig:riemanngeomech}(c), the curvature is positive at $\jointangle_{2}=0$ and falls away for non-zero values of $\jointangle_{2}$ as values of $\mixedconn$ begin to saturate. Consequently, for a set of geometrically-similar gaits, displacement per cycle grows subquadratically with the scale of the gait (quadratically increasing area with a diminishing average integrand), as illustrated in \zcref[S]{fig:riemanngeomech}(b), and matching the description of the gaits in \zcref[S]{exp:gaitdirectintegration}.

Because of the relative areas they enclose, displacements induced by elliptical gaits are approximately $\frac{\pi}{4}$ the displacements induced by same-size box gaits (completely shifting the mass before expanding or contracting the blocks). At larger amplitudes, the displacements induced by the box and elliptical gaits become slightly closer; geometrically, this property corresponds to the corner-sections enclosed by the boxes but not by the ellipses being biased towards $\jointangle_{2}$ values where $D\mixedconn$ is small and making a relatively small contribution to the total displacement induced by the box gaits.
\end{example}

A full analysis and optimization of system gait performance requires not only characterizing the net displacement induced by gaits, but also the costs (in terms of time or energy) incurred while executing the gait---an easy ``stride" that can be repeated quickly will generally produce faster motion at a given effort than will a long ``lunge" that maximizes use of the range of shape motion, but takes longer to complete each cycle. Details of this cost and efficiency structure are provided in~\citet{Ramasamy:2019aa}; we do not review them here because our focus in this paper is on generalizing models for system displacement from systems with physics defined by Riemannian metrics to those whose physics are defined by Finsler metrics, and save the question of efficiency for future work.

\section{Finsler mechanical systems}

As discussed above, linear-viscous drag models inherently have the mathematical structure of a Riemannian metric, which makes them amenable to differential-geometric analysis, producing the constraint-curvature and path length cost insights described in \S\ref{sec:gaits}.
This mathematical congruence prompts the questions ``Are there other environmental-contact models for which we can produce similar geometric insights'', and, in particular, ``What aspects of the geometric locomotion structure remain if we relax some of the conditions on the friction model?''

For the remainder of this paper, we examine the consequences of replacing the Riemannian metric that encodes linear-viscous friction with a \emph{Finsler metric} encoding homogeneous-viscous friction. Intuitively, this new model preserves the convexity and  proportionality of the linear-viscous model---such that, e.g., doubling the velocity at a contact point doubles the resulting resistance force, and the resistance interpolating between velocities dissipates at least as much power as either endpoint---but lifts two other conditions on linearity: reversibility and linear additivity.

Removing reversibility allows us to consider ``ratchet-like'' systems which have large drag when pushed in one direction and small drag when pushed in the opposite direction---e.g. systems with angled scales or bristles that ``dig in'' or ``fold out of the way'' depending on the direction of the contact slip. Similarly, removing linear additivity allows us to consider systems whose contact structure couples the drag in different directions such that it cannot be decomposed into ``forward" and ``lateral" components---e.g. because of buckling elements in the contact interface.

In this section, we begin the generalization of our locomotion framework from Riemannian to Finsler structure by first examining the similarities and differences between the two classes of metrics and consider mathematical representations of Finsler metrics. We then discuss the construction of a sub-Finslerian constraint on the system motion analogous to the sub-Riemannian constraint used in conventional geometric locomotion analysis~\cite{Montgomery:2002vn}. This construction lays the groundwork for the following section, in which we generalize our notion of constraint curvature to the sub-Finslerian geometry and identify a new component of this curvature that enables locomotion modes not available to systems operating under linear-viscous friction laws.

\subsection{From Riemann to Finsler}

Riemannian metrics $\metric$ define norms
\begin{equation}
\norm{\configdot}_{\metric} = \sqrt{\transpose{\configdot} \metric(\config)\,  \configdot},
\end{equation}
for vectors tangent to a manifold at each point $\config$. These norms satisfy the standard positive-definite and triangle-inequality requirements for norms,
\begin{equation}
\norm{\configdot} > 0.
\end{equation}
and
\begin{equation}
\norm{\configdot_{1} + \configdot_{2}} \leq \norm{\configdot_{1}} + \norm{\configdot_{2}},
\end{equation}
and are additionally \emph{positive-homogeneous}, such that scaling a vector $v$ by a factor $\lambda \geq 0$ scales the norm by the same factor,
\begin{equation}
\norm{\lambda v} = \lambda \norm{v}.
\end{equation}
The homogeneous property means that the graph of the norm function is a cone (all level sets have the same shape, are concentric, and are evenly spaced), and the bilinear structure under the square root means that the level sets are ellipsoids centered on the origin.
\begin{figure}
\centering
\includegraphics[width=0.5\linewidth]{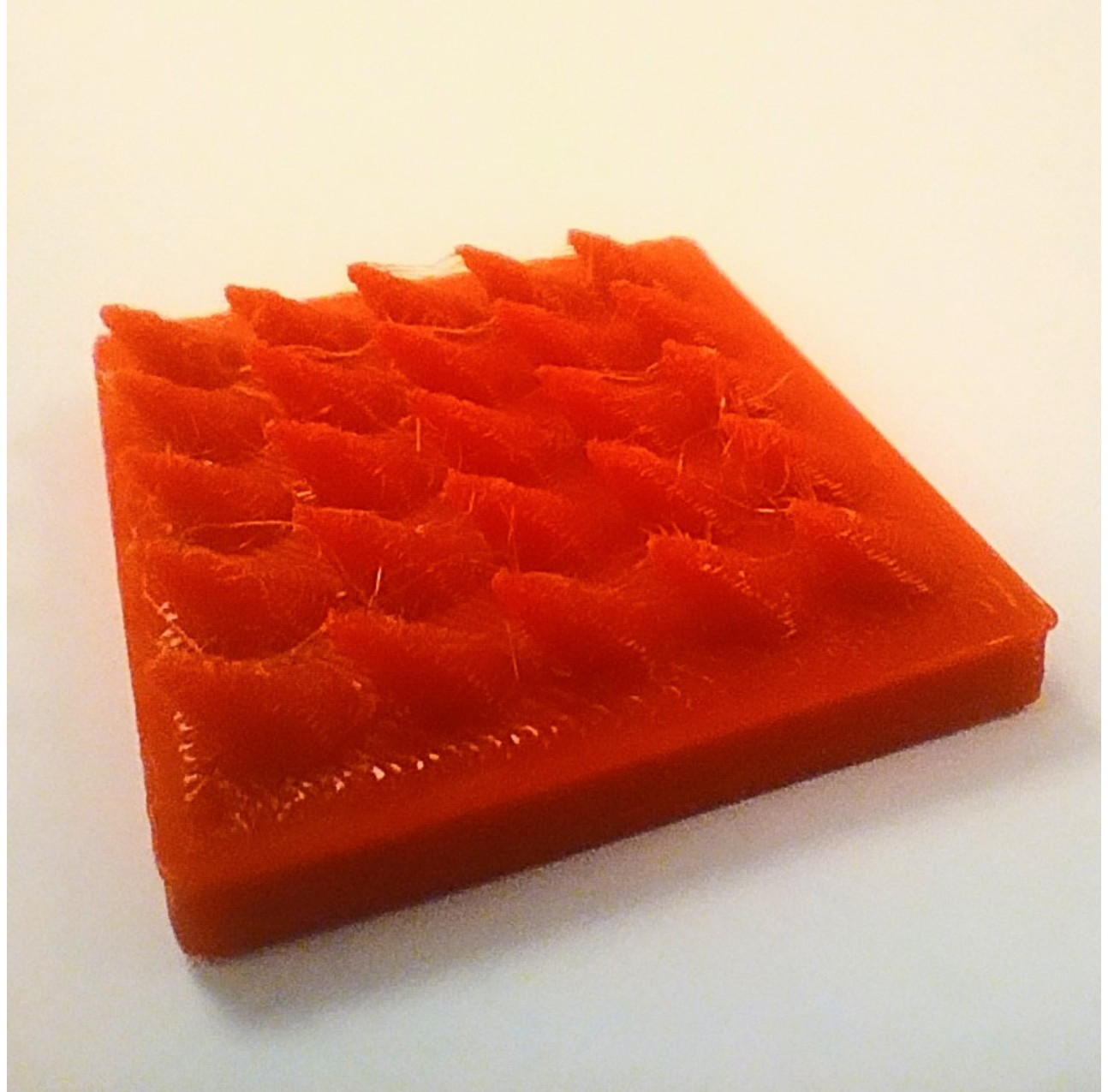}
\caption{Example of spines generating a asymmetric friction force. %
When applied to low ply industrial carpet, these spines have many times the friction sliding in one direction vs. the opposite direction.}
\label{fig:spines}
\end{figure}

\begin{example} \label{exp:singleblockriemanfriction}
A block sliding on a line with viscous friction coefficient $\fricco$ has a Riemannian norm
\begin{equation}
\norm{\xdot}_{M} = \sqrt{P} = \sqrt{\fricco\xdot^{2}} = \sqrt{\fricco} \, \abs{\xdot}
\end{equation}
As illustrated in \zcref[S]{fig:basicfinsler}(a), the graph of this function is V-shaped (a cone over one-dimensional space), with a slope proportional to the square root of the viscous friction coefficient. Its level sets (pairs of points with the same value) are centered on the origin and evenly spaced.
\end{example}

\begin{figure}
\centering
\includegraphics[width=\linewidth]{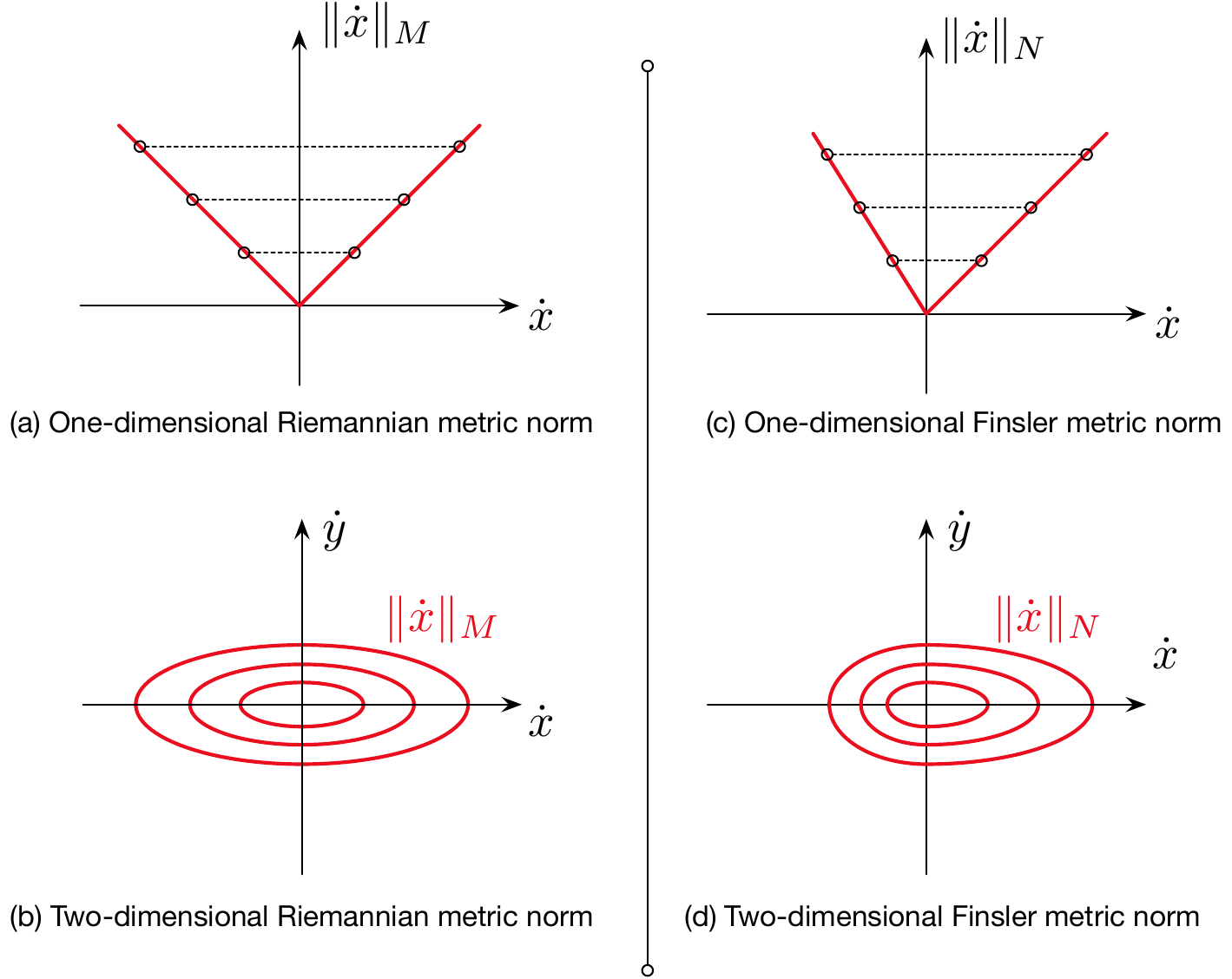}
\caption{Geometry of vector norms under Riemannian and Finsler metrics. %
(a) Riemannian metric norms are positive-homogeneous (scaling the vector scales the norm proportionally), and their level sets are centered around the origin. %
(b) In more than one dimension, the level sets of the metric are ellipsoidal, but remain centered around the origin. %
(c) Finsler metric norms are positive-homogeneous, but their level sets do not need to be symmetric around the origin. %
(d) In more than one dimension, Finsler metric norms are functions whose level sets are convex, concentric, evenly spaced, and enclose the origin. %
The illustrated metric is the union of two half-ellipses, and could plausibly represent the friction generated by the spines in \zcref[S]{fig:spines}. %
See \zcref[S]{exp:singleblockriemanfriction, exp:singleblockriemanfriction2d} for mathematical representations of the Riemannian metrics, and \zcref[S]{exp:finsler1dsingleblock,exp:finsler2dsingleblock} for representations of the Finsler metrics.}
\label{fig:basicfinsler}
\end{figure}

\begin{example} \label{exp:singleblockriemanfriction2d}
A block sliding on a plane with different viscous coefficients $\fricco_{x}$ and $\fricco_{y}$ in two orthogonal directions has a Riemannian norm
\begin{equation}
\norm{v}_{\metric} = \sqrt{p} = \sqrt{\begin{bmatrix} \xdot & \ydot \end{bmatrix} \begin{bmatrix}
\fricco_{x} & 0 \\0 & \fricco_{y}
\end{bmatrix}
\begin{bmatrix} \xdot \\ \ydot \end{bmatrix}}
\end{equation}
As illustrated in \zcref[S]{fig:basicfinsler}(b), the graph of this function is a cone with elliptical level sets centered on the origin. The long axes of these ellipses are aligned with the direction of \emph{smaller} drag: the cone is shallower along this axis, and thus it takes a larger velocity component in that direction to produce a given norm value.
\end{example}

Finsler metrics $\Finmetric$ are generalizations of Riemannian metrics in which the norm function
\begin{equation}
\norm{\configdot}_{\Finmetric} = \Finmetric(\config,\configdot)
\end{equation}
can be any function satisfying the positive-homogeneous, positive-definite, and triangle-inequality properties. As in the Riemannian case, the homogeneity and definite properties mean that the graph of the norm function is a cone; the triangle-inequality property means that the level sets must be convex curves enclosing the origin (with the centered ellipses of a Riemannian metric a special case of this condition).

\begin{example} \label{exp:finsler1dsingleblock}
A block sliding on a line with viscous coefficients $\fricco^{+}$  in the forward direction and $\fricco^{-}$ in the backward direction has a Finsler norm
\begin{equation}
\norm{\xdot}_{\Finmetric} = \begin{cases}
\fricco^{+}\, \abs{\xdot} & \xdot\geq0 \\
\fricco^{-}\, \abs{\xdot} & \xdot<0
\end{cases} .
\end{equation}
As illustrated in \zcref[S]{fig:basicfinsler}(c), the graph of this function is an asymmetric V, steeper in the direction of with the larger coefficient. Unlike the Riemannian case, the level-set point pairs are not centered on the origin.
\end{example}
\begin{example} \label{exp:finsler2dsingleblock}
A block sliding on a plane with viscous coefficients $\fricco_{x}^{+}$ and $\fricco_{x}^{-}$ for forward and backward motion and $\fricco_{y}$ for both lateral directions has a Finsler norm
\begin{equation}
\norm{v}_{\Finmetric} = \sqrt{P} = \begin{cases}
\sqrt{\left[\begin{smallmatrix} \xdot & \ydot \end{smallmatrix}\right] \left[\begin{smallmatrix}
c_{x}^{+} & 0 \\0 & c_{y}
\end{smallmatrix}\right]
\left[\begin{smallmatrix} \xdot \\ \ydot \end{smallmatrix}\right]}
& \xdot\geq0\\[1.5ex]
\sqrt{\left[\begin{smallmatrix} \xdot & \ydot \end{smallmatrix}\right] \left[\begin{smallmatrix}
c_{x}^{-} & 0 \\0 & c_{y}
\end{smallmatrix}\right]
\left[\begin{smallmatrix} \xdot \\ \ydot \end{smallmatrix}\right]} & \xdot < 0
\end{cases}
\end{equation}
As illustrated in \zcref[S]{fig:basicfinsler}(d), the graph of this function is a cone whose level sets are seed-shaped, formed piecewise from two-half ellipses.
\end{example}

\begin{figure}
\centering
\includegraphics[width=0.5\linewidth]{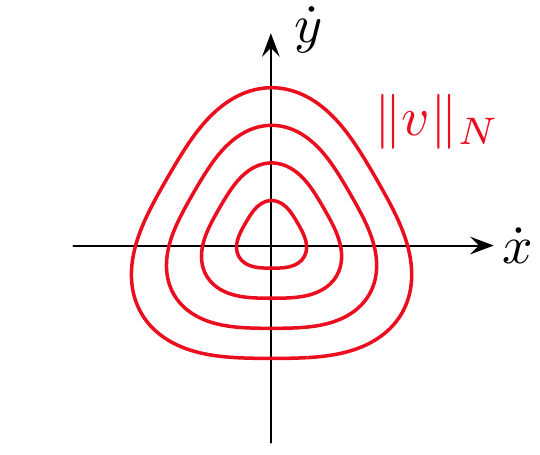}
\caption{Finsler metric defined by a polar function. In contrast to the Finsler metric defined by a piecewise union of Riemannian metrics in  \zcref[S]{fig:basicfinsler}(d), this metric function (from \zcref[S]{exp:polarfinsler}) has been constructed directly from a polar function, and designed to exhibit trilateral symmetry.}
\label{fig:polarfinsler}
\end{figure}

\begin{example} \label{exp:polarfinsler}
The Finsler metrics discussed in~\zcref[S]{exp:finsler1dsingleblock,exp:finsler2dsingleblock} are both constructed piecewise from Riemannian metrics.
However, consider an alternative to the pair of ellipses:
\begin{equation}
\norm{v}_{\Finmetric} = |v| (9+\sin(3\measuredangle v)),
\end{equation}
where $|v|$ refers to the $\mathcal{L}_2$ norm.
This function has level sets that are convex, rounded, and trilaterally symmetric shapes, as illustrated in \zcref[S]{fig:polarfinsler}.
The friction is not symmetric---for example, motion in the $+y$ direction produces $5/4$ more drag power than motion in the $-y$ direction.
\end{example}

\subsection{System-level Finsler metrics}
Given a set of metrics defining drag at the system's contact points with the world, we can construct a metric by pulling back the local metrics through the system kinematics: For a set of Finsler metrics $\Finmetric_{i}$ at the contact points, the aggregate metric on the configuration space is
\begin{equation} \label{finslersystempower}
\Finmetric(\config,\configdot) = \sum_{i} \Finmetric_{i}(\config, \jac_{i}(\config) \configdot),
\end{equation}
such that the norm of a configuration velocity is equal to the sum of the norms of the contact velocities it induces.
Both the property of being positively homogeneous and the property of being convex are closed under composition on linear maps (from Lemma~\ref{lem:posHomOfLin} and Lemma~\ref{lem:conCompAffine}, respectively; see Appendix~\ref{sec:apxA}), and under finite sums (from Lemma~\ref{lem:sumPosHom} and Lemma~\ref{lem:convexSum}; see Appendix~\ref{sec:apxA}).

The pullback of the local Riemannian drag in~\eqref{eq:Riemannpullback} is a special case of this operation in which the $\configdot$ term naturally factors out of the calculation; generalizing the pullback operation to a function composition (instead of matrix operations) allows for the construction of system metrics where, as in \zcref[S]{exp:polarfinsler}, the local metrics are not (piecewise) Riemannian.

\begin{figure}
\centering
\includegraphics[width=0.9\linewidth]{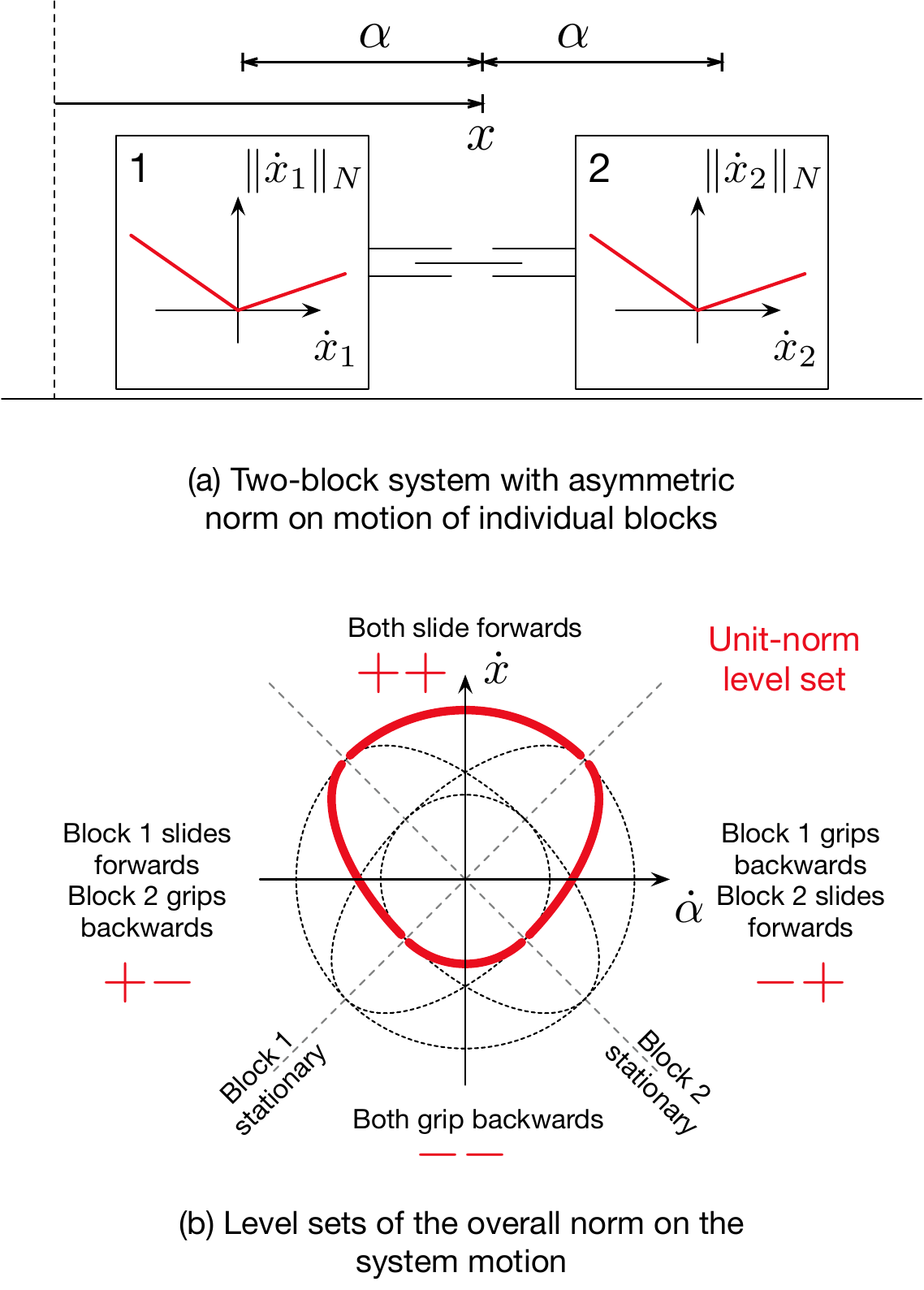}
\caption{A two-block system with asymmetric friction has four possible forward/backward contact states, based on the ratio and direction of the system bulk motion relative to its shape change motion. The level sets of the overall system Finsler norm are thus a piecewise joining of the level sets of the Riemannian norms for the corresponding contact states, each of which is a circle or an ellipse.}
\label{fig:twoblockfinslerlevelset}
\end{figure}

\begin{figure*}
\centering
\includegraphics[width=\linewidth]{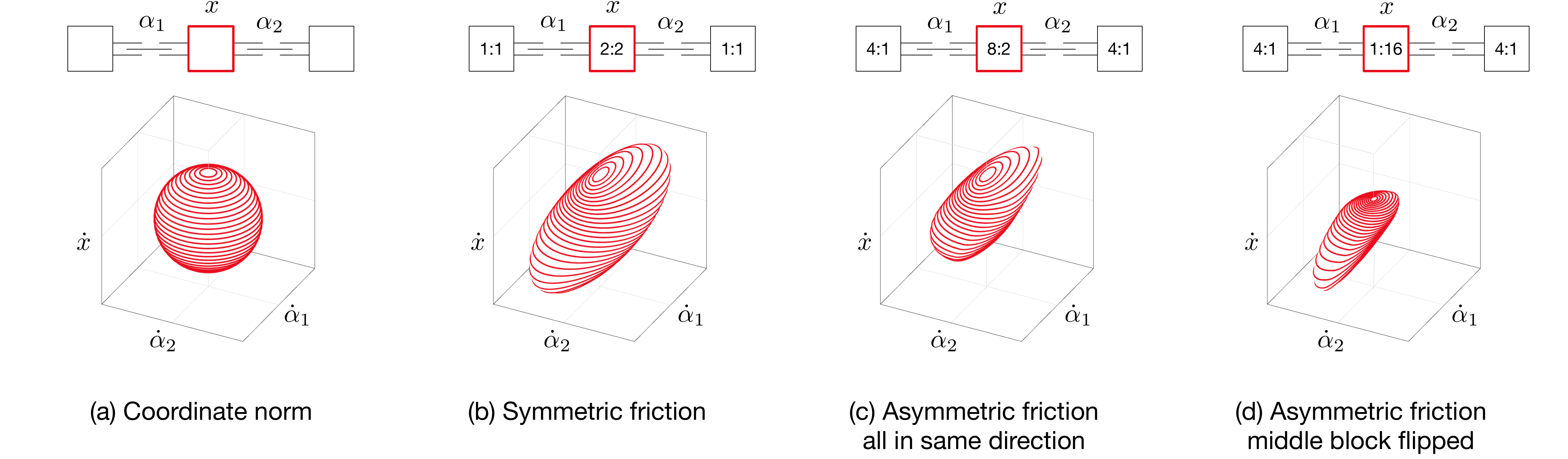}
\caption{Metric level sets for a three-block system, as described in \zcref[S]{exp:threeblock}. We generate the level sets by starting with the unit sphere of coordinate velocities illustrated in (a), and then for each system metric evaluate the norm at each point on the sphere and divide that point by its norm to place it on the metric level set. Position is taken as that of the middle block, and the blocks are subject to the backwards:forwards friction coefficients labeled inside the blocks, with (b) symmetric friction, (c) asymmetric friction penalizing backwards motion of each block, and (d) asymmetric friction penalizing backwards motion of the outer blocks and forwards motion of the middle block.}
\label{fig:three-block-level-sets}
\end{figure*}

\begin{example} \label{exp:twoblockfinsler}
The system in \zcref[S]{fig:twoblockfinslerlevelset}(a)
consists of two blocks connected by a linear actuator (i.e., it has the same mechanical structure as the planarized skier in \S\ref{sec:geomechoverview}, but with the weight removed or held in the center). As before, its position $\fiber=(x)$ is the mean of the two block positions, and its shape $\jointangle$ is half the distance between the blocks. The Jacobians for the motion of the two blocks are again
\begin{equation}
\dot{x}_{1} = \overset{\displaystyle{\jac_{1}}}{\begin{bmatrix} 1 & -1 \end{bmatrix}}\begin{bmatrix}
\bodyvel \\ \jointangledot
\end{bmatrix}
\qquad
\text{and}
\qquad
\dot{x}_{2} = \overset{\displaystyle{\jac_{2}}}{\begin{bmatrix} 1 & 1 \end{bmatrix}}\begin{bmatrix}
\bodyvel \\ \jointangledot
\end{bmatrix},
\end{equation}
with positive body motions moving both blocks forwards and positive shape changes moving both blocks outwards.

If we subject the blocks to the Riemannian friction metrics described in \zcref[S]{exp:singleblockriemanfriction}, the system metric is pulled back to a metric whose level sets are circles,
\begin{equation}
\dragmetric = \begin{bmatrix} \pmi 1 \\ -1 \end{bmatrix} [c] \begin{bmatrix} 1 & -1 \end{bmatrix} + \begin{bmatrix} 1 \\ 1 \end{bmatrix} [c] \begin{bmatrix} 1 & 1 \end{bmatrix} = \begin{bmatrix} 2c & 0 \\ 0 & 2c \end{bmatrix}.
\end{equation}
If we instead apply directional friction to the blocks as in \zcref[S]{exp:finsler1dsingleblock}, with $c^{+}=c$ and $c^{-} = 4c$, each block contributes two possible metric tensors to the system metric (depending on its direction of motion),
\begin{subequations}
\begin{align}
\Findragmetric_{1+} &= \begin{bmatrix} \pmi c & -c \\ -c & \pmi c \end{bmatrix}
&&&
\Findragmetric_{2+} &= \begin{bmatrix} c & c \\ c & c \end{bmatrix}
\\
\Findragmetric_{1-} &= \begin{bmatrix} \pmi 4c & -4c \\ -4c & \pmi 4c \end{bmatrix}
&&&
\Findragmetric_{2-} &= \begin{bmatrix} 4c & 4c \\ 4c & 4c \end{bmatrix},\\
\intertext{which combine as} \nextParentEquation
\Findragmetric_{++} &= \begin{bmatrix} 2c & 0 \\ 0 & 2c \end{bmatrix}
&&&
\Findragmetric_{+-} &= \begin{bmatrix}  5c & 3c \\ 3c &  4c \end{bmatrix}
\\
\Findragmetric_{-+} &=  \begin{bmatrix} \pmi 5c & -3c \\ -3c & \pmi 5c \end{bmatrix}
&&&
\Findragmetric_{--} &= \begin{bmatrix}  8c & 0 \\ 0 & 8c \end{bmatrix}.
\end{align}
\end{subequations}

The unit-value level sets of the metric norms corresponding to these four tensors are illustrated in \zcref[S]{fig:twoblockfinslerlevelset}(b). The $\Findragmetric_{++}$ and $\Findragmetric_{--}$ norms both have circular level sets (with the greater magnitude of the $\Findragmetric_{--}$ coefficients producing a \emph{smaller} circle), and the level sets of the $\Findragmetric_{+-}$ and $\Findragmetric_{-+}$ norms are angled ellipses.

The aggregate Finsler norm for the system velocity can be constructed from these norms by piecewise-composition, using the Jacobians from the $(\jointangle, \xdot)$ space to the individual block motions to determine which metric is ``active''. For this system, $\Findragmetric_{++}$ is active when the system $\xdot$ body motion is positive enough to overcome any $\jointangledot$-induced backwards motion of block 1, and $\Findragmetric_{--}$ is active when $\xdot$ is negative enough to overcome $\jointangledot$-induced forwards motion of block 2. The level-set arc segments corresponding to these conditions are joined by the ellipse segments corresponding to the cases where the $\jointangledot$-induced expansion or contraction dominates the bulk $\xdot$ motion, producing a rounded geometry which is convex but asymmetric in the $\xdot$ direction.
\end{example}

\begin{example}\label{exp:threeblock}

The three-block systems illustrated in \zcref[S]{fig:three-block-level-sets} provide examples of system-level Finsler metrics for systems with two shape variables. As in the two-block example, we take the $\jointangle$ values as the half-distance between blocks; we take the position of this system as the location of the middle block.

With symmetric friction on each block, the metric is Riemannian, and its level sets are ellipsoids centered on the origin. Moving the actuators opposite to each other while translating center block produces the least individual block motion (because the outer blocks stay essentially unmoving while the middle block shuffles between them), so the level set is longest in this direction.

Placing asymmetric friction on the blocks makes the system metric Finslerian, with backwards motion more expensive than forward motion, and biases the level sets towards positive-$\xdot$ velocities. Flipping the direction of the friction asymmetry on the middle block makes forwards motion more expensive, biasing the level sets towards negative-$\xdot$ values.
\end{example}

\subsection{Sub-Finsler Motility}
As discussed in \S\ref{sec:subriemannconds}, the motility map for a shape-actuated system takes input shape velocities $\jointangledot$ to the system body velocities $\bodyvel$ they induce. For a drag-dominated system, these $\bodyvel$ induced by a given $\jointangledot$ can be considered either as that for which the position-component of the combined drag on $\begin{bmatrix} \bodyvel & \jointangledot \end{bmatrix}$ is zero (i.e., for which the individual contact-drag forces are in equilibrium around the system body frame), or equivalently, as those which minimize the norm of $\begin{bmatrix} \bodyvel & \jointangledot \end{bmatrix}$. In previous contexts with linear-viscous friction, this constraint has been referred to as sub-Riemannian (as the system motion is restricted to a subspace of the tangent bundle of a Riemannian manifold); in the context of our current study, this minimization becomes a sub-Finslerian condition on the system motion.

The power-minimizing points are located at points where the derivative of the norm with respect to position velocity goes to zero,
\begin{equation} \label{subfinslerconds}
\frac{\partial P}{\partial \bodyvel} = 0.
\end{equation}
Geometrically, these points are located at points where the tangents of the level sets of the norm function are aligned with the position directions.\footnote{More specifically, at points where projecting any velocity vector with only position components into the tangent directions does not change the vector.} The convexity of the Finsler norm level sets (which is preserved through the linear pullback operations encoded by the Jacobians) guarantees that there is a unique power-minimizing position velocity for each shape velocity, and that there are no local maxima which would also satisfy the zero-derivative condition.

\begin{figure*}
\centering
\includegraphics[width=0.95\linewidth]{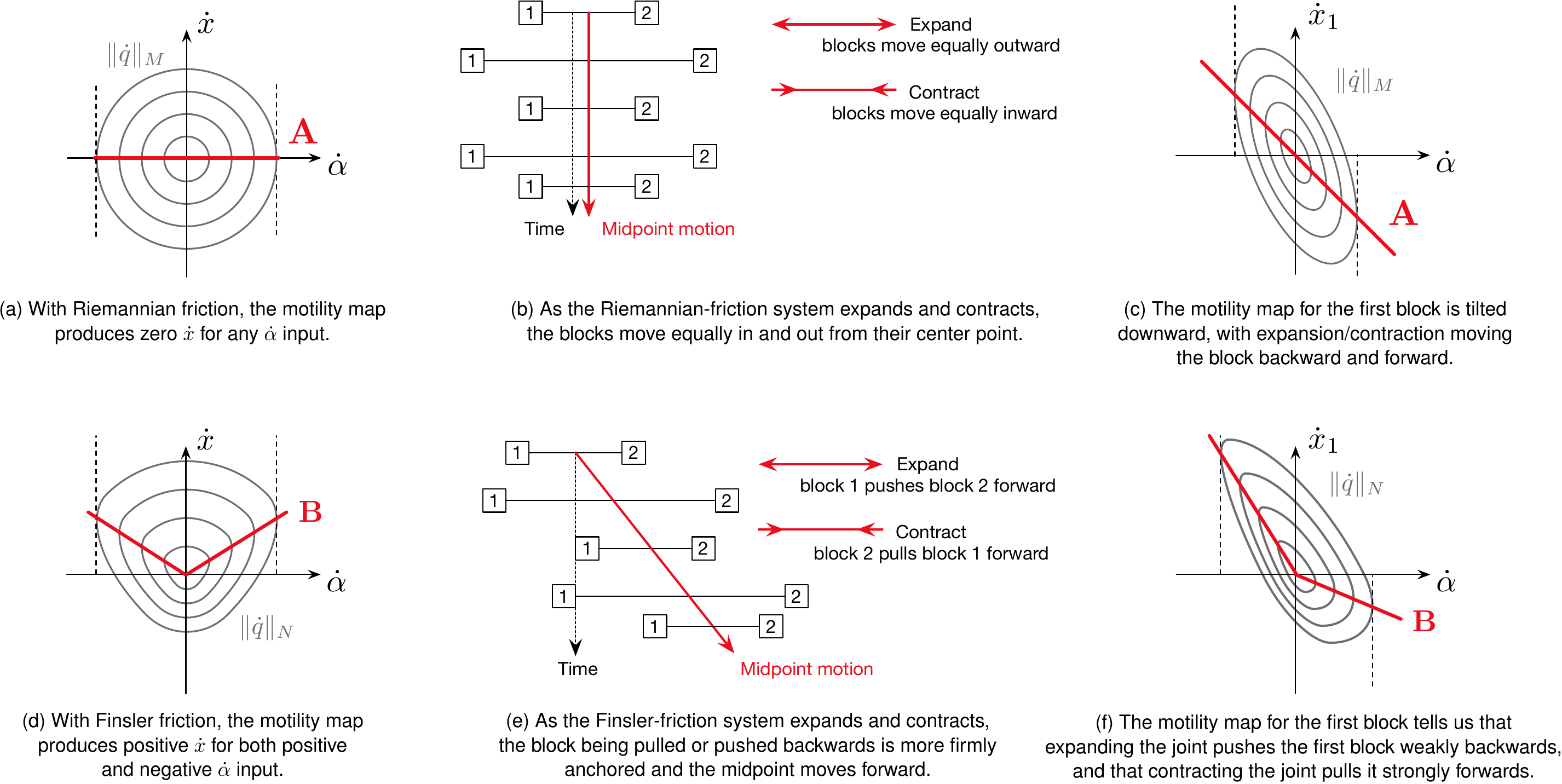}
\caption{Motility maps for a two-block system with (a)--(c) symmetric Riemannian friction, as described in \zcref[S]{exp:twoblockriemannmotility}, and (d)--(f) asymmetric Finsler friction, as described in \zcref[S]{exp:twoblockfinslermotility}. In (b) and (d), the horizontal axis is the $x$ axis of the system and the vertical axis is time.
}
\label{fig:twoblockmotility}
\end{figure*}

\begin{example}\label{exp:twoblockriemannmotility}
If we apply symmetric (Riemannian) friction to the two-block system illustrated in \zcref[S]{fig:twoblockfinslerlevelset}, the norm level sets are circles as illustrated in \zcref[S]{fig:twoblockmotility}(a), such that the $\frac{\partial P}{\partial \bodyvel} = 0$ points are located at $\xdot=0$, with a motility map
\begin{equation}
\mixedconn = -\inv{\metric}_{\fiber\fiber}\metric_{\fiber\jointangle} = -\begin{bmatrix}\frac{1}{2c}\end{bmatrix} \begin{bmatrix} 0 \end{bmatrix}= \begin{bmatrix} 0 \end{bmatrix}.
\end{equation}
Oscillating the shape of the system pushes the blocks evenly out and in around the midpoint with symmetric reaction forces, so the midpoint does not move, as illustrated in \zcref[S]{fig:twoblockmotility}(b).

Changing coordinates so that the position of the system is that of the first block (i.e., setting $\fiber = x_{1}$) makes the expression for the metric tensor
\begin{equation}
\dragmetric = \overbrace{\begin{bmatrix} 1 \\ 0 \end{bmatrix} [c] \begin{bmatrix} 1 & 0 \end{bmatrix}}^{\textrm{block 1}} + \overbrace{\begin{bmatrix} 1 \\ 2 \end{bmatrix} [c] \begin{bmatrix} 1 & 2 \end{bmatrix}}^{\textrm{block 2}} = \begin{bmatrix} 2c & 2c \\ 2c & 4c \end{bmatrix},
\end{equation}
where the Jacobians indicate that in these coordinates the motion of the first block is independent of the joint velocity, and that the joint velocity now contributes twice to the second block velocity (center point relative to first block, and then second block relative to center point.

The level sets of the metric in these coordinates are tilted-ellipses as illustrated in \zcref[S]{fig:twoblockmotility}(c). The $\frac{\partial P}{\partial \bodyvel} = 0$ points are then located at $\xdot=-\jointangledot$, with a motility map
\begin{equation}
\mixedconn = -\inv{\metric}_{\fiber\fiber}\metric_{\fiber\jointangle} = -\begin{bmatrix}\frac{1}{2c}\end{bmatrix} \begin{bmatrix} \frac{1}{2c} \end{bmatrix}= \begin{bmatrix} - 1 \end{bmatrix}.
\end{equation}
This motility map encodes the behavior observed in \zcref[S]{fig:twoblockmotility}(b) that expanding the joint pushes the first block backwards and contracting the joint draws it forwards, with both motions at the same rate, and thus producing no net displacement.
\end{example}

\begin{figure*}
\centering
\includegraphics[width=0.85\linewidth]{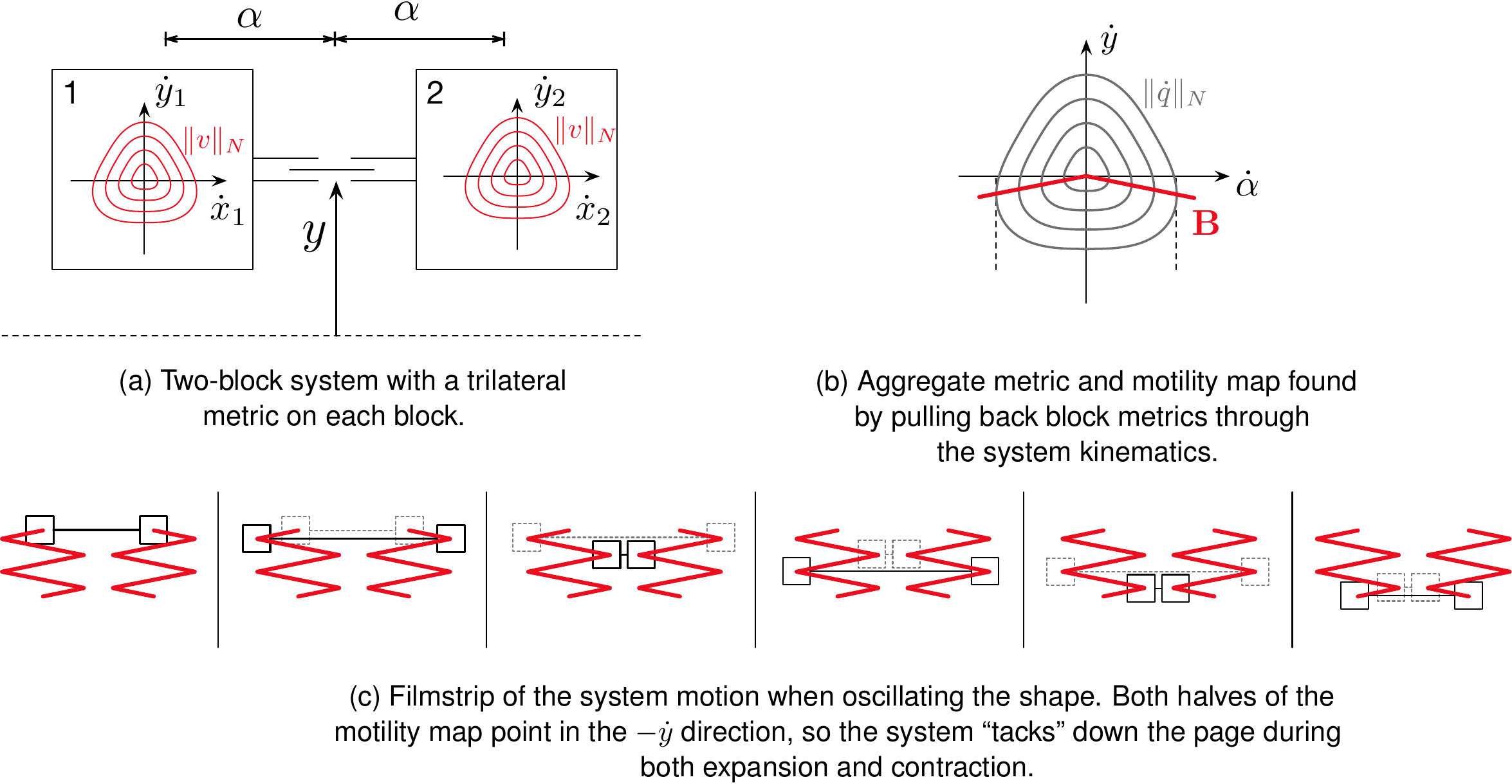}
\caption{Motility map for a two-block system with trilateral Finsler metric friction on each link. %
(a) System configuration and orientation of the individual friction metrics on each block. %
(b) These Finsler metrics pull back to a body frame, and this defines a motility map through finding the points where friction does minimal work. %
(c) As a consequence, moving the blocks back and forth generates a transverse motion downward.}
\label{fig:twoblocktrilateralfriction}
\end{figure*}

\begin{figure*}
\centering
\includegraphics[width=0.95\linewidth]{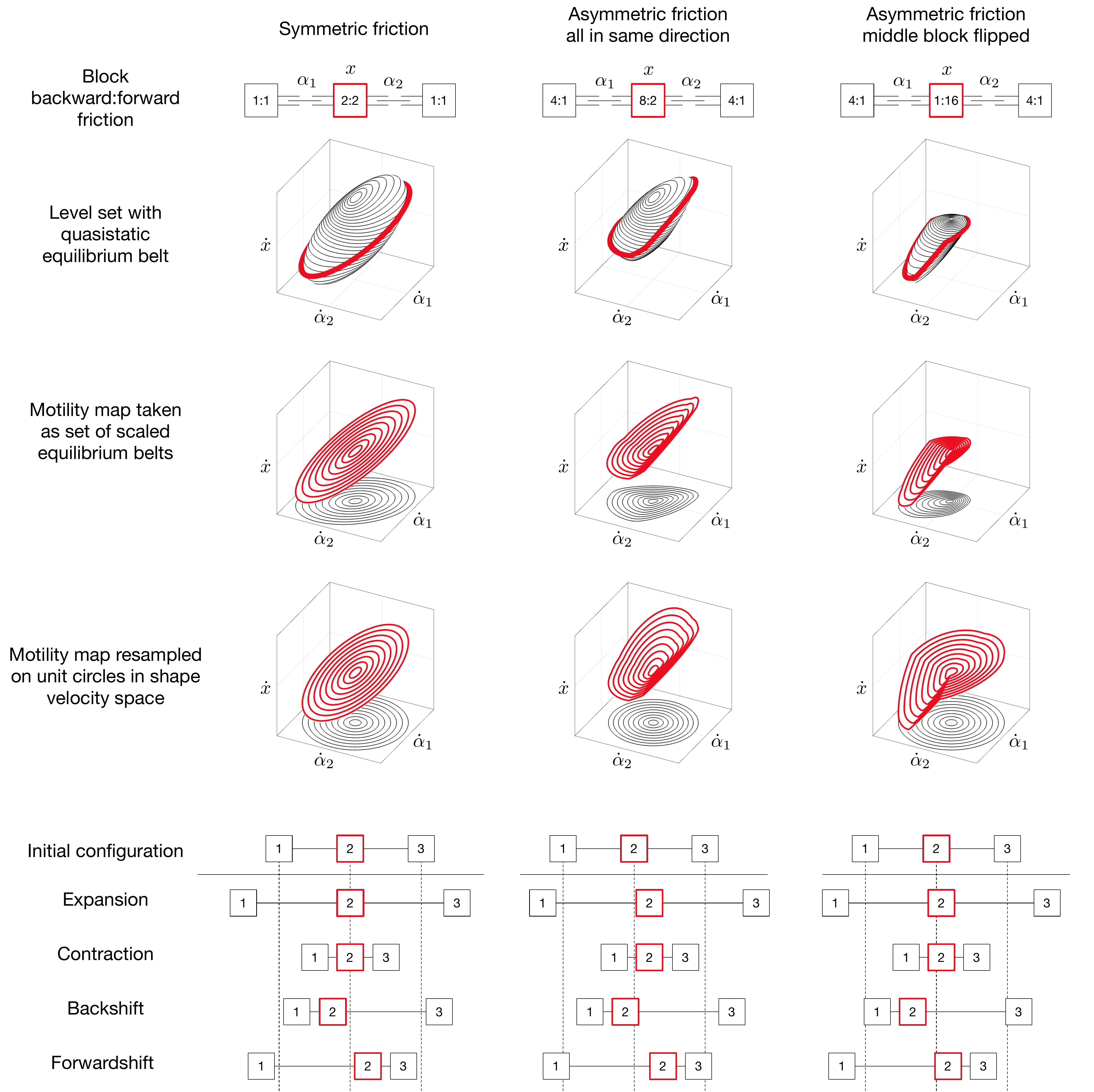}
\caption{Motility maps for three-block systems with different friction arrangements on the blocks, as described in \zcref[S]{exp:threeblockmotility}. The motility map for the symmetric-friction (i.e., Riemannian) system in the first column is linear, with expansion/contraction of the joints producing no motion, and back- or forwardshifting the middle block relative to the outer blocks producing equal amounts of motion. Placing directed friction on the blocks, as in the second column, forward-biases all of the motions  and flipping the directed friction on the middle block as in the third column retains the forward bias on expansion-contraction motions, but introduces a \emph{negative} bias for back- and forwardshifting the middle block. These biases can be seen both in structure of the motility map functions and by comparing the post-shape-change block positions in the bottom part of the figure.}
\label{fig:threeblockmotility}
\end{figure*}

\begin{example}\label{exp:twoblockfinslermotility}
Using the asymmetric friction model discussed in \zcref[S]{exp:twoblockfinsler} and illustrated in \zcref[S]{fig:twoblockfinslerlevelset} makes the level sets of the power norm seed-shaped. As illustrated in \zcref[S]{fig:twoblockmotility}(d), the $\frac{\partial P}{\partial \bodyvel} = 0$ points for this system are in the regions where the$\Findragmetric_{-+}$ and $\Findragmetric_{+-}$ metrics are active, such that the motility map is
\begin{equation}
\finconn =
\begin{cases}
-\inv{\Finmetric}_{\frac{-+}{\fiber\fiber}}\Finmetric_{\frac{-+}{\fiber\jointangle}} = -\begin{bmatrix} 5c \end{bmatrix}\begin{bmatrix} -3c \end{bmatrix} = \begin{bmatrix} \pmi \frac{3}{5} \end{bmatrix} & \jointangledot \geq 0
\\
-\inv{\Finmetric}_{\frac{+-}{\fiber\fiber}}\Finmetric_{\frac{+-}{\fiber\jointangle}} = -\begin{bmatrix} 5c \end{bmatrix}\begin{bmatrix} \pmi 3c \end{bmatrix} = \begin{bmatrix} -\frac{3}{5} \end{bmatrix} & \jointangledot < 0
\end{cases},
\end{equation}
and the relationship between the joint and the midpoint can be succintly expressed as $\xdot=\frac{3}{5}\abs{\jointangledot}$.
The absolute value in this expression reflects the property that, as illustrated in  \zcref[S]{fig:twoblockmotility}(e), both expanding and contracting the system moves its center of mass forwards: during expansion, the backwards motion of block 1 generates a large gripping force that pushes block 2 forward against its smaller forward friction; during contraction, the roles of the blocks are reversed, with block 2 providing an anchoring force to pull block 1 forward.

The mirror symmetry in the norm and motility map for this system reflects the property that the joint moves the two blocks symmetrically with respect to the midpoint. If we change coordinates to make the first block the system location, the level sets of the metric expression become warped as illustrated in \zcref[S]{fig:twoblockmotility}(f). The expressions of this metric under the $-+$ and $+-$ sliding conditions become
\begin{align}
\Findragmetric_{-+} &= \begin{bmatrix}  5c & 1c \\ 2c &  4c \end{bmatrix}
&&&
\Findragmetric_{+-} &=  \begin{bmatrix} 5c & 8c \\8c & 16 c \end{bmatrix},
\end{align}
such that the motility map for the first block evaluates to
\begin{equation}
\finconn =
\begin{cases}
-\inv{\Finmetric}_{\frac{-+}{\fiber\fiber}}\Finmetric_{\frac{-+}{\fiber\jointangle}} = -\begin{bmatrix} 5c \end{bmatrix}\begin{bmatrix} 2c \end{bmatrix} = \begin{bmatrix} - \frac{2}{5} \end{bmatrix} & \jointangledot \geq 0
\\
-\inv{\Finmetric}_{\frac{+-}{\fiber\fiber}}\Finmetric_{\frac{+-}{\fiber\jointangle}} = -\begin{bmatrix} 5c \end{bmatrix}\begin{bmatrix} 8c \end{bmatrix} = \begin{bmatrix} -\frac{8}{5} \end{bmatrix} & \jointangledot < 0
\end{cases}.
\end{equation}
Both portions of $\finconn$ have negative slopes, but the $\jointangledot \geq 0$ portion has a shallower slope. These properties correspond to the system motion illustrated in \zcref[S]{fig:twoblockmotility}(f), in which expanding the joint moves the first block backwards by a small amount, and contracting it moves the block forward by a large amount.
\end{example}

\begin{example}
If we apply the trilaterally symmetric friction model from \zcref[S]{exp:polarfinsler} to the blocks, as illustrated in \zcref[S]{fig:twoblocktrilateralfriction}(a), the aggregate metric for motion in the $y$ and $\jointangle$ directions is also trilaterally symmetric, as illustrated in \zcref[S]{fig:twoblocktrilateralfriction}(b).\footnote{Symmetries in the system mean that the system experiences no $x$ motion, so we can focus on only the $y$ motion} The vertical tangents to the level sets of this metric are both at negative $\dot{y}$ values, meaning that the system moves in the $-y$ direction under both expansion and contraction of the joint, as illustrated in \zcref[S]{fig:twoblocktrilateralfriction}(c).
\end{example}

\begin{example}\label{exp:threeblockmotility}
The motility maps for the three-block systems in \zcref[S]{exp:threeblock} are constructed from the set of all points in $(\jointangledot_{1}, \jointangledot_{2}, \xdot)$ where the tangent plane to the system's level set at that point is vertical (i.e., where the normal vector to the level set is normal to $\xdot$).
These points represent minimum-power/force-equilibrium velocities, and form a ``belt" around the level set, as illustrated in the second row of \zcref[S]{fig:threeblockmotility}.

Because the level sets of a Finsler metric are by definition convex, these points can be found by the following steps, illustrated in the third row of \zcref[S]{fig:threeblockmotility}:
\begin{enumerate}
\item Projecting a representative level set into the $(\jointangledot_{1}, \jointangledot_{2})$ plane (black curves).
\item Finding the convex hull of this projection.
\item Restoring the $\dot{x}$ values to the hull points (one red curve).
\item Scaling the resulting points proportionally to find the belts that lie on other level sets (all red curves).
\end{enumerate}
Once this set of points has been constructed, a more balanced representation of it can be constructed by interpolating into it with a set of concentric rings in the shape space, as illustrated in the fourth row of \zcref[S]{fig:threeblockmotility}.

As illustrated in the first column of \zcref[S]{fig:threeblockmotility}, the motility map for the system with symmetric friction is a plane, much as the motility map for the two-block system with symmetric friction was a line. In both cases, this linearity follows from the Riemannian nature of the symmetric friction.

The plane has zero slope along the $\jointangledot_{1} = \jointangledot_{2}$ line. Physically, this property corresponds to the symmetric friction on the outer blocks acting symmetrically on the middle block during expansion or contraction of the system, as illustrated at the bottom left of \zcref[S]{fig:threeblockmotility}. Its steepest slope is along the $\jointangledot_{1} = -\jointangledot_{2}$ line, where shifting the middle block toward the first or last block generates pulling and forces between the blocks. The middle block moves the same amount as each of the outer blocks because its friction coefficient is equal to the sum of the friction coefficients on the other two blocks.

Putting asymmetric friction on the blocks causes the motility map to become cupped, as illustrated in the second column of \zcref[S]{fig:threeblockmotility}. This cupping corresponds to the property that both expansion and contraction along the $\jointangledot_{1} = \jointangledot_{2}$ shape axis involve one of the outer blocks acting as an anchor (as in the two-block case), with the middle block additionally providing less resistance to the outer blocks when it is sliding forward. than when it is sliding backwards.

For motion along the $\jointangledot_{1} = -\jointangledot_{2}$ axis, the asymmetric friction means that the middle block acts as an anchor when shifted towards the first block, giving the motility map a shallow slope in the $(-\jointangledot_{1}, +\jointangledot_{2})$ direction, but glides easily when shifted towards the last block, making the motility map steep in the $(+\jointangledot_{1}, -\jointangledot_{2})$ direction.

If we flip the direction of the asymmetry, as in the third column of \zcref[S]{fig:threeblockmotility}, the slopes in the $\jointangledot_{1} = \jointangledot_{2}$ become less pronounced, because the opposing asymmetries partially cancel each other, making the system behave more like the symmetric-friction system under expansion and contraction.

In the $\jointangledot_{1} = -\jointangledot_{2}$ axis, flipping the sign of the middle asymmetry means that backshifting the middle block becomes easier, steepening the $(-\jointangledot_{1}, +\jointangledot_{2})$ slope, but that forwardshifting it becomes harder, making the $(+\jointangledot_{1}, -\jointangledot_{2})$ slope more shallow. This change in slopes has the effect of changing the motility map function from the ``tilted cone" geometry in the same-direction system to a ``tilted saddle" in the flipped-direction case.
\end{example}

\subsection{Low-order Approximation of Sub-Finslerian Motility.} \label{sec:linearconicaldecomposition}

\begin{figure}
\centering
\includegraphics[width=0.45\textwidth]{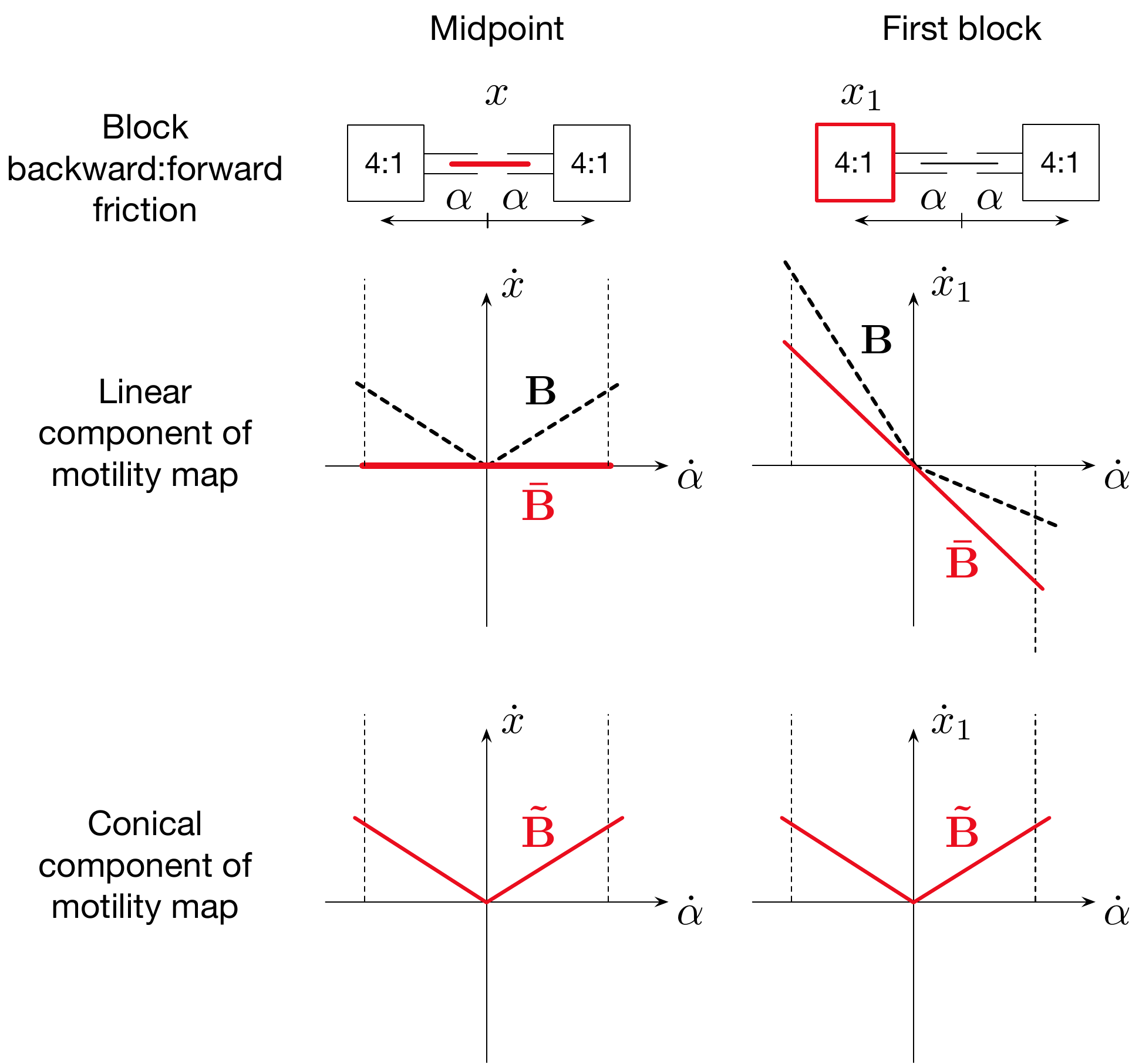}
\caption{Linear-conical decomposition of the motility maps for the two-block system with two choices of position coordinate, as discussed in \zcref[S]{exp:twoblockmotilityapprox}. The mean slopes $\finconnmean$ for the two-block system with asymmetric friction, whose motility map was found in \zcref[S]{exp:twoblockfinslermotility}, are the lines through the origin that best fit $\finconn$ at $\jointangledot=\pm 1$, and the conical components are the residuals, which are equal to each other.}%
\label{fig:twoblockmotilityapproximation}
\end{figure}

\begin{figure*}
\centering
\includegraphics[width=\textwidth]{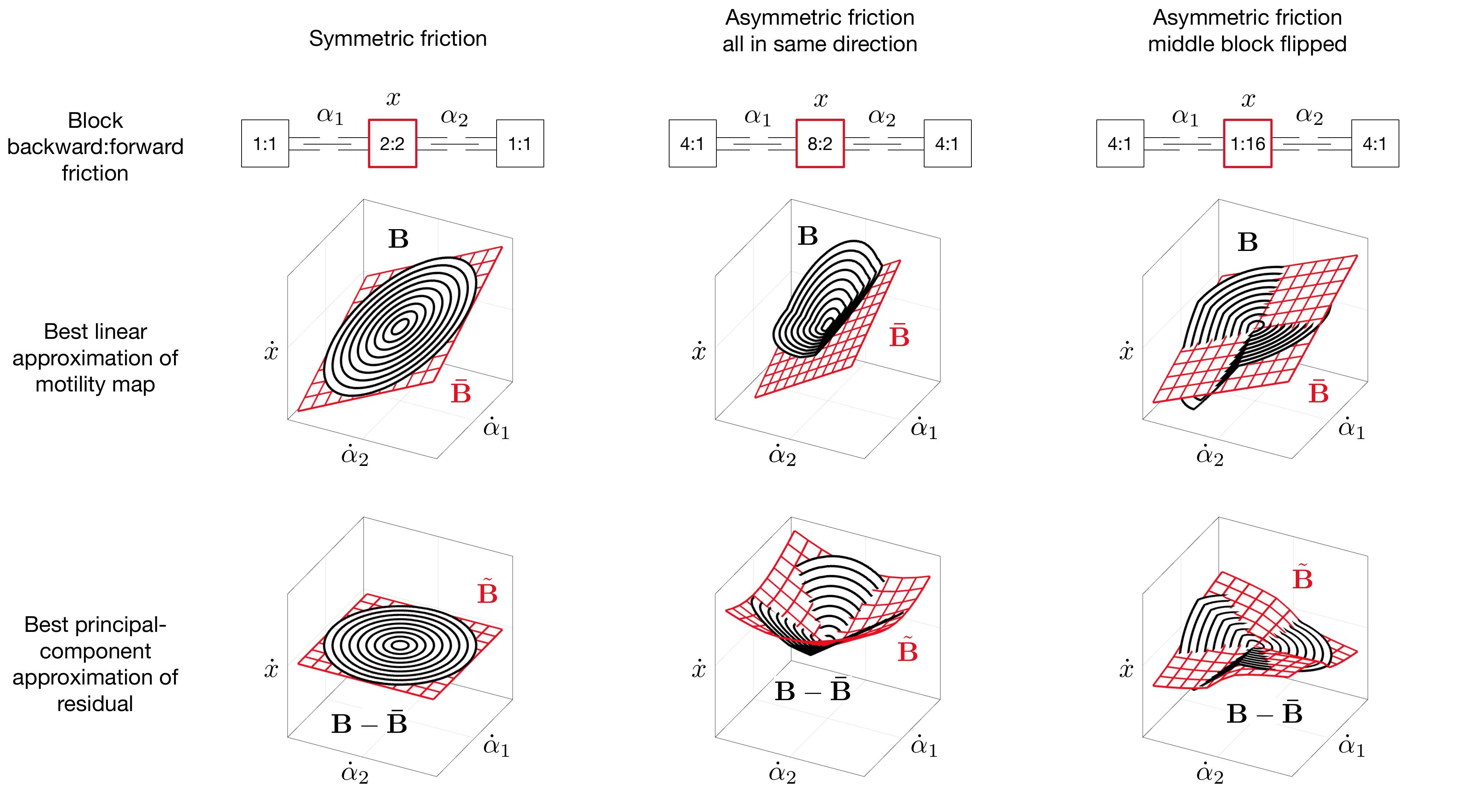}
\caption{Linear-conical approximation of the motility map for the three-block systems. The mean slopes $\finconnmean$ are the planes whose mean-squared differences from $\finconn$ along the unit circle in $\jointangledot$ are smallest, and the conical components $\finconndiff$ are the generalized cones that best fit the residual portion. These cones may be flat (if $\finconn$ is entirely captured by $\finconnmean$, as in the symmetric-friction case at left), bowl-shaped with an elliptical cross-section (if the principal components of the cone are both of the same sign, as in the asymmetric case at center), or saddle-shaped (if the principal components are of opposite signs, as in the middle-flipped case at right).}
\label{fig:threeblockmotilityapproximation}
\end{figure*}

One of the benefits of the sub-Riemannian locomotion paradigm is that it describes the system motility at any given point in terms of a small number of intuitively-describable parameters: the slope of position with respect to shape in each shape direction. This simplicity of construction enables high-level reasoning and analysis about system properties, and in particular about the constraint curvature which determines the net motion achievable via gaits.

Although sub-Finslerian motility maps are in general more complicated functions of shape velocity, we can extend many of the benefits of a low-dimensional representation to such systems by approximating sub-Finslerian motility maps as the sum of a linear component $\finconnmean$ and a generalized conical/saddle component $\finconndiff$,
\begin{equation} \label{eq:finconndecomposition}
\finconn(\jointangle,\jointangledot) \approx \finconnmean(\jointangle,\jointangledot) + \finconndiff(\jointangle,\jointangledot) + \ldots,
\end{equation}
which at each $\jointangle$ respectively capture the principal asymmetric and symmetric aspects of $\finconn(\jointangledot)$.

In this decomposition, $\finconnmean$ is a standard linear form defined by its slope in each shape direction,
\begin{align}
\finconnmean(\jointangle,\jointangledot) &= \textstyle{\sum_{i}} \finconnmean_{i}(\jointangle)\, \jointangledot_{i},
\\
\intertext{and $\finconndiff$ is an elliptical cone or saddle constructed as the sign-preserving square root of a quadratic function,}
\finconndiff(\jointangle,\jointangledot) &= \left(\textstyle{\sum_{i,j}} \mathbf{K}_{ij}(\jointangle)\, \jointangledot_{i} \jointangledot_{j}\right)^{\star\frac{1}{2}},
\\
\intertext{which can equivalently be described in terms of a set of principal magnitudes and directions,}
\finconndiff(\jointangle,\jointangledot) &=\left( \textstyle{\sum_{i}} \Bigl( \finconndiff_{j}^{\vphantom{T}}(\jointangle) \Bigr)^{\star 2}
\Bigl( \transpose{V}_{j}(\jointangle)\cdot \jointangledot\Bigr)^{2}\right)^{\star\frac{1}{2}}, \label{eq:Btildedef}
\end{align}
where we again use the $\star$ symbol to denote sign-preserving power operations.

The $\finconnmean_{i}$ principal linear components and $(\finconndiff_{j},V_{j})$ principal conical components best approximating a sub-Finslerian motility map can be calculated at a given $\jointangle$ by sampling $\finconn$ along an even sampling of unit shape velocities $\hat{\jointangledot}$ in the tangent space at that point and performing a set of linear regressions and eigen-decompositions against those points. Details of this process are discussed in \zcref[S]{app:subRapprox}. If continuity of the approximation across multiple $\jointangle$ values is desired, the fitting operation can be extended to fitting analytical functions and their derivatives across regions of the shape space.%

\begin{example} \label{exp:twoblockmotilityapprox}
The linear and conical components of the motility maps for the two-block (single shape degree of freedom) system with asymmetric friction from \zcref[S]{exp:twoblockfinslermotility} are illustrated in \zcref[S]{fig:twoblockmotilityapproximation}. The linear components $\finconnmean$ are the lines whose differences from $\finconn$ at $\jointangledot=\pm 1$ are of equal magnitude, and the conical portions $\finconndiff$ are (as for all systems with a single shape variable) exactly the residual portion once the mean has been removed.

The mean slopes for the motility of the midpoint and first block are different, but their conical portions are equal. This equality of $\finconndiff$ across changes of position coordinates is a fundamental property across all sub-Finslerian systems, including those with more than one shape variable: changes in shape coordinate add a linear term to the motility (corresponding to the Jacobian of the coordinate change), and this linear term is captured by $\finconnmean$.
\end{example}

\begin{example} \label{exp:threeblockmotilityaprox}

The linear and conical components of the motility maps for the three-block systems from \zcref[S]{exp:threeblockmotility} are illustrated in \zcref[S]{fig:threeblockmotilityapproximation}. For all three systems, $\finconnmean$ is in the $(+\jointangle_{1}, -\jointangle_{2})$ direction, corresponding to the property that moving the middle block away from block 1 and towards block 3 slides it forwards with respect to the world under all the friction regimes. Beyond this shared property, the different friction regimes generate different second-order structures:
\begin{itemize}
\item The system with symmetric friction is sub-Riemannian, so its motility map is completely captured by $\finconnmean$, and its $\finconndiff$ component is zero. %
\item The system with asymmetric friction in the same direction on each block has a residual $\finconndiff$ component that is %
a level bowl shape which can be approximated by an elliptical cone, whose principal axes are aligned with the $(+\jointangle_{1}, -\jointangle_{2})$ and $(+\jointangle_{1}, +\jointangle_{2})$ directions.
\item The system with the asymmetric friction flipped on the middle block %
has a saddle-shaped residual, reflecting the property that both $(+\jointangle_{1}, +\jointangle_{2})$ and $(-\jointangle_{1}, -\jointangle_{2})$ shape changes push the system forward, and that $(-\jointangle_{1}, +\jointangle_{2})$ shape changes push the system backwards more than $(+\jointangle_{1}, -\jointangle_{2})$ push it forwards.

The small residuals left after fitting the regularized cone and saddle functions $\finconndiff$ to $\finconn - \finconnmean$ have approximately trilateral symmetry, corresponding to the next term in the decomposition series in \zcref[S]{eq:finconndecomposition} (and thus the leading error in the linear-conical approximation) being a cubic-generated homogeneous function (the cube root of a cubic polynomial).
\qedhere
\end{itemize}%
\end{example}

\section{Approximating Net Motion for sub-Finslerian Systems}

The constraint curvature relationship between gait geometry and net motion for sub-Riemannian systems in~\eqref{eq:constraintcurvatureintegral},
\begin{equation} \label{eq:constraintcurvatureintegralredux}
\fiberexp_{\gait} \approx \iint_{\gait} \overbrace{\extd \mixedconn + \liebracket{\mixedconn_{i}}{\mixedconn_{j>i}}}^{D\mixedconn}.
\end{equation}
can be generalized to sub-Finslerian systems by applying the linear-conical decomposition from \S\ref{sec:linearconicaldecomposition}:
\begin{enumerate}
\item Calculating the nonconservative and noncommutative curvature of the motility map using the $\finconnmean$ linear component of the sub-Finsler motility map in place of the $\mixedconn$ sub-Riemannian motility map.
\item Introducing a new ``nonreciprocal" curvature component, $\finconndiff$, that provides a low-order description of the changes in the motility map that appear when reversing the direction of motion.
\item Providing approximation rules for the integral of $\finconndiff$, enabling enabling reasoning about the average (and thus net) motion induced by nonreciprocal effects during a gait to $\nonrecip\finconn$ and the gait's geometry.
\end{enumerate}

The $\finconnmean$ linear component of a sub-Finslerian motility map has the same structure as a sub-Riemannian motility map, and captures the ``bias" of $\finconn$ at each shape. The nonconservative curvature of this bias, $\extd\finconnmean$, describes the derivative of the bias with respect to system shape such that the induced motion does not completely cancel out over a gait cycle. Similarly, the bias's noncommutative curvature, $\liebracket{\finconnmean_{i}}{\finconnmean_{j>i}}$, captures position-space interactions between vectors in $\finconnmean$ that lead to global net motion over a cycle, even if the local expressions cancel. %

\subsection{Constructing the Nonreciprocal  Constraint Curvature}

In addition to nonconservative motility changes across the shape space and noncommutative changes across the position space, sub-Finslerian systems can also exhibit \emph{nonreciprocal} motility changes associated with reversing the direction of shape velocity: If the slope of $\finconn$ at a given shape is different in the $+\jointangledot$ and $-\jointangledot$ directions, oscillating through that shape will produce a non-zero average velocity. In the same spirit as $\extd\finconnmean$ and $\liebracket{\finconnmean_{i}}{\finconnmean_{j>i}}$ locally capture the nonconservative and noncommutative effects as a curvature of the system constraints, we can define a nonreciprocal curvature term $\nonrecip\finconn$ that locally captures these cross-direction changes and is approximately equal to
the conical component $\finconndiff$ constructed in \S\ref{sec:linearconicaldecomposition}:
\begin{subequations}
\begin{align}
\nonrecip\finconn(\jointangle, \jointangledot) &= \textstyle{\frac{1}{2}}\bigl(\finconn(\jointangle,\jointangledot) - (-\finconn(\jointangle,-\jointangledot))\bigr)
\\
&=\textstyle{\frac{1}{2}}\bigl(\finconn(\jointangle,\jointangledot) + \finconn(\jointangle,-\jointangledot) \bigr)
\\
&\approx \finconndiff(\jointangle, \jointangledot),
\end{align}
\end{subequations}
(where the approximation is accurate up to the extent that $\finconndiff$ approximates $\finconn-\finconnmean$).

This nonreciprocal curvature behaves somewhat differently from the nonconservative and noncommutative curvatures. Firstly, unlike those curvatures, this new curvature is calculated via a subtraction operation instead of a derivative; this property corresponds to reversing shape velocity being a discrete operation, whereas the other two terms are evaluated with respect to continuous changes in configuration.\footnote{Intuitively, this construction is analogous to the way in which vertex angles replace surface curvatures in the geometry of discrete objects.}

Secondly, $\nonrecip\finconn(\jointangle,\jointangledot)$ does not describe the infinitesimal contribution to average motion from a gait that encircles a given region, but instead the infinitesimal contribution from a gait that oscillates through $\jointangle$ in the $\pm\jointangledot$ direction, with
\begin{subequations}
\begin{align}
\bodyvel_{\text{mean}}(\jointangle,\jointangledot) &= \nonrecip\finconn(\jointangle,\jointangledot) + \nonrecip\finconn(\jointangle,-\jointangledot)
\\
&=2\nonrecip\finconn(\jointangle,\jointangledot)
\\
&=2\finconndiff(\jointangle,\jointangledot).
\end{align}
\end{subequations}

\begin{figure*}
\centering
\includegraphics[width=\textwidth]{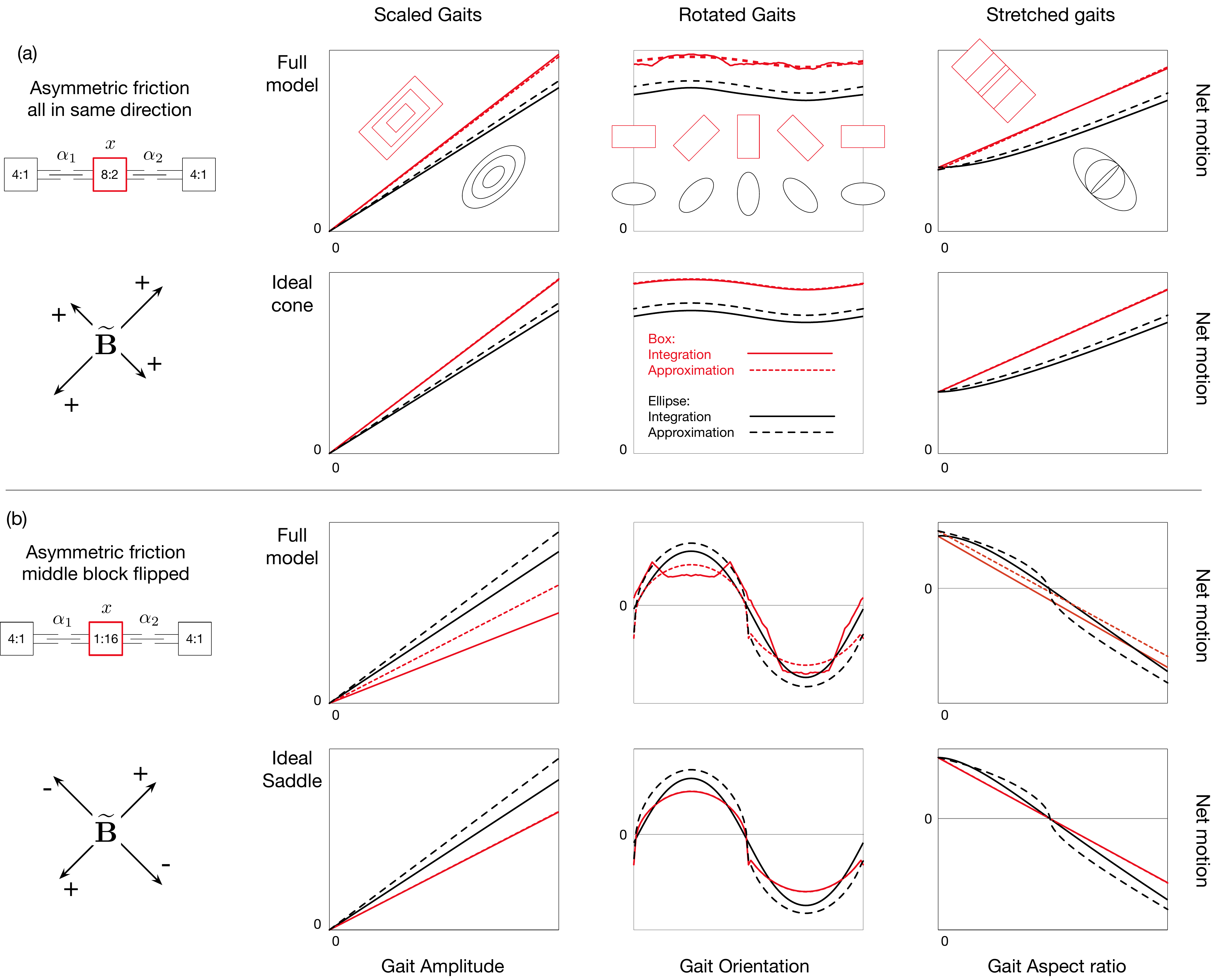}
\caption{Net motion approximations for three-block systems with (a) asymmetric friction on the blocks, all in the sme direction, and (b) asymmetric friction on the blocks, with the middle block biased in the opposite direction. For each system, we plot the net motion induced by box- or ellipse-shaped gaits with different scales, orientations, and aspect ratios, and compare these net motions with the predictions based on the $\finconndiff$ principal components of the nonreciprocal curvature. For each system, we plot the same values for an ``ideal'' system constructed from the $\finconndiff$ cone or saddle function, highlighting which errors come from the gait-evaluation approximations, and which come from the use of a low-order characterization of the system dynamics. In all plots, solid lines indicate the ``true" integrated motion of the system, dashed lines indicate the box- or ellipse- approximations, black lines are for box-shaped gaits, and red lines are for ellipse-shaped gaits.}
\label{fig:threeblockapproximations}
\end{figure*}
\subsection{Integrating Nonreciprocal Constraint Curvature}

Over a gait, the net average motion is approximately equal to the integral of the standard curvature of $\finconnmean$ plus the integral of $\finconn-\finconnmean$ along the gait,
\begin{equation}
\fiberexp_\gait \approx \iint_{\gait} D\finconnmean + \int_{\gait} (\finconn-\finconnmean).
\end{equation}
As discussed in \zcref[S]{app:BCH}, we have not included Lie bracket interactions associated with $\finconn-\finconnmean$ in this expression, and will consider their contribution in future work.

To provide a tractable approximation for understanding the non-reciprocal contribution to the net displacement, we first note that first-order variations in $\finconn-\finconnmean$ do not contribute to its integral over a gait, such that we can reasonably approximate $(\finconn-\finconnmean)$ at the $\jointangle_{0}$ center of the gait, rather than at every point along it. This single-point evaluation then allows us to replace the difference with the non-reciprocal curvature at the center point, and then further approximate it with the conical component, such that
\begin{subequations}
\begin{align}
\int_{\gait} (\finconn-\finconnmean) &\approx \int_{\gait} (\finconn_{\jointangle_{0}}-\finconnmean_{\jointangle_{0}}) \\&\approx \int_{\gait} \nonrecip\finconn_{\jointangle_{0}} \approx \int_{\gait} \finconndiff_{\jointangle_{0}}.
\end{align}
\end{subequations}

Based on this approximation, and in particular the symmetry of $\nonrecip\finconn$ and $\finconndiff$ with respect to $\jointangledot$, we can make two broad statements about the expected contribution from nonreciprocal effects to the net motion:
\begin{enumerate}
\item The nonreciprocal contribution scales linearly with the size of the gait (as compared to quadratically for the nonconservative and noncommutative contributions).
\item The sign of the nonreciprocal contribution is independent of the clockwise/counterclockwise orientation with which the gait was executed (whereas flipping the orientation of a gait flips the signs of the nonconservative and noncommutative contributions).
\end{enumerate}
The specific relationship between gait geometry, nonreciprocal curvature geometry, and net induced motion is somewhat more complicated to specify, but at a high level, the induced displacement is a product of the principle components of the gait and the curvature, modulated by the alignment between their respective principal axes.
These attributes combine according to the ``roundness" of the gait, with some useful cases including:
\begin{enumerate}
\item For simple one-dimensional gaits in which the shape traces back and forth along a line defined by a direction $\hat{\gait}$ and a length $\ell$, the net induced motion is twice the length of the gait multiplied by the nonreciprocal curvature in its direction,
\begin{equation}
\fiberexp_{\nonrecip} \approx 2\ell\finconndiff(\jointangle_{0}, \hat{\gait}).
\end{equation}
\item For rectangular gaits defined by an orientation basis $\hat{\gait}$ and a set of sidelengths $\ell$ (or more generally, polygonal gaits with antiparallel opposing sides of the same length), the net induced motion is the sum of the motion induced by the simple linear gaits in the corresponding directions,
\begin{equation} \label{eq:rectangularapproximation}
\fiberexp_{\nonrecip} \approx {\textstyle \sum_{j}}2\ell_{j}\finconndiff(\jointangle_{0}, \hat{\gait}_{j}).
\end{equation}
\item For elliptical gaits, the expected net induced motion can be approximated by treating the gait as a regular octagon, aligned with and stretched by the ellipse's diameters. Using the general rule above for polygonal gaits, this approximation resolves to a weighted sum of the simple and signed-$L^{2}$ sums of the axis-aligned simple gaits,
\begin{multline} \label{eq:ellipticalapproximation}
\fiberexp_{\nonrecip} \approx  \frac{1}{1+\sqrt{2}} {\textstyle \sum_{j}}2\ell_{j}\finconndiff(\jointangle_{0}, \hat{\gait}_{j})
\\
+ \frac{\sqrt{2}}{1+\sqrt{2}} \Bigl({\textstyle \sum_{j}}\bigl(2\ell_{j}\finconndiff(\jointangle_{0}, \hat{\gait}_{j})\bigr)^{\star2}\Bigr)^{\star\frac{1}{2}}.
\end{multline}
Because such an octagon exscribes the elliptical gait path, the magnitude of the predicted displacement can be expected to slightly overpredict the magnitude of the net displacement.

\end{enumerate}
If second-order changes in $\finconndiff$ across the shape space are significant, these approximations can be improved by pushing the evaluation of $\finconndiff$ out to the edges of the gait. For example, in a polygonal gait, if the segments of the $j$th pair of sides are centered at $\jointangle_{j}^{+}$ and $\jointangle_{j}^{-}$, then we can make an improved approximation via the substitution
\begin{equation}
2\finconndiff(\jointangle_{0}, \hat{\gait}_{j}) \Longrightarrow \finconndiff(\jointangle_{j}^{+}, \hat{\gait}_{j}) + \finconndiff(\jointangle_{j}^{-}, \hat{\gait}_{j})
\end{equation}
into \zcref[S]{eq:rectangularapproximation, eq:ellipticalapproximation}.

\begin{example}
The two-block systems have only a single shape variable, and so (because the exterior derivative and Lie bracket are zero or undefined in spaces with fewer than two dimensions) can only locomote through non-reciprocal motility contributions. For these systems, $\finconndiff$ is simply the difference between the $\jointangledot^{+}$ and $\jointangledot^{-}$ slopes of the motility map, and multiplying the length of a gait's span in the shape space by this difference exactly produces the time-normalized average velocity,
\begin{equation}
\fiberexp_{\gait} = L \finconndiff.
\end{equation}
\end{example}

\begin{example}
The motility maps for the three-block systems are constant functions over their shape spaces, meaning that their $\finconnmean$ mean slopes are constant, and thus that their $\extd\finconnmean$ nonconservative curvatures are zero. Consequently, the net motions experienced by the systems over gait cycles can be attributed to their $\nonrecip\finconn$ nonreciprocal curvatures.

As illustrated in \zcref[S]{fig:threeblockapproximations}, the net displacement induced by box- or ellipse-shaped gaits scales linearly with the amplitudes of the gaits. For the system with same-direction asymmetric friction on all blocks, illustrated in \zcref[S]{fig:threeblockapproximations}(a), the  scaling rates are closely matched by their respective $\finconndiff$-based approximations. The accuracy of the predictions holds both for gaits rotated away from the $\finconndiff$ principal axes, whose net motion changes as the major axis of the gait becomes aligned with the larger or smaller $\finconndiff$ principal components, and for gaits of different aspect ratios, in which the gait starts aligned with one component, and then increasingly adds a term from the second component.

The system with opposite-direction friction on the middle block, illustrated in \zcref[S]{fig:threeblockapproximations}(b) shows a larger rate-discrepancy for box gaits than for either the ellipse gaits or the same-direction system. This discrepancy does not appear if we reconstruct an ``ideal saddle" $\nonrecip\finconn$ from $\finconndiff$, indicating that the source of the error is in the difference between our low-order approximation and the true system dynamics. This attribution is further corroborated by the sets of rotated and stretched gaits, where the box approximation has error in the full model, but smaller error in the reconstructed ideal saddle.

The ellipse-gait approximations show somewhat more error for the opposite-direction friction system than for the same-direction system, but this error is more consistent than the box-gait approximations on this system. This consistency is largely due to the ellipse gaits averaging their contribution to the net motion over all directions, whereas the box gaits pick out just two directions, and are thus more sensitive to differences between the low-order and full models.

We also note that the relative prediction errors for the opposite-direction system are significantly larger than the relative errors between $\nonrecip\finconn$ and $\finconndiff$: Because we are essentially subtracting the contribution from one principal axis from that of the other, the ``ground truth" value for the net motion is smaller than the magnitude of $\nonrecip \finconn$, but the absolute errors from using $\finconndiff$ to approximate the motion are not reduced in scale.
\end{example}

\begin{figure*}
\centering
\includegraphics[width=\textwidth]{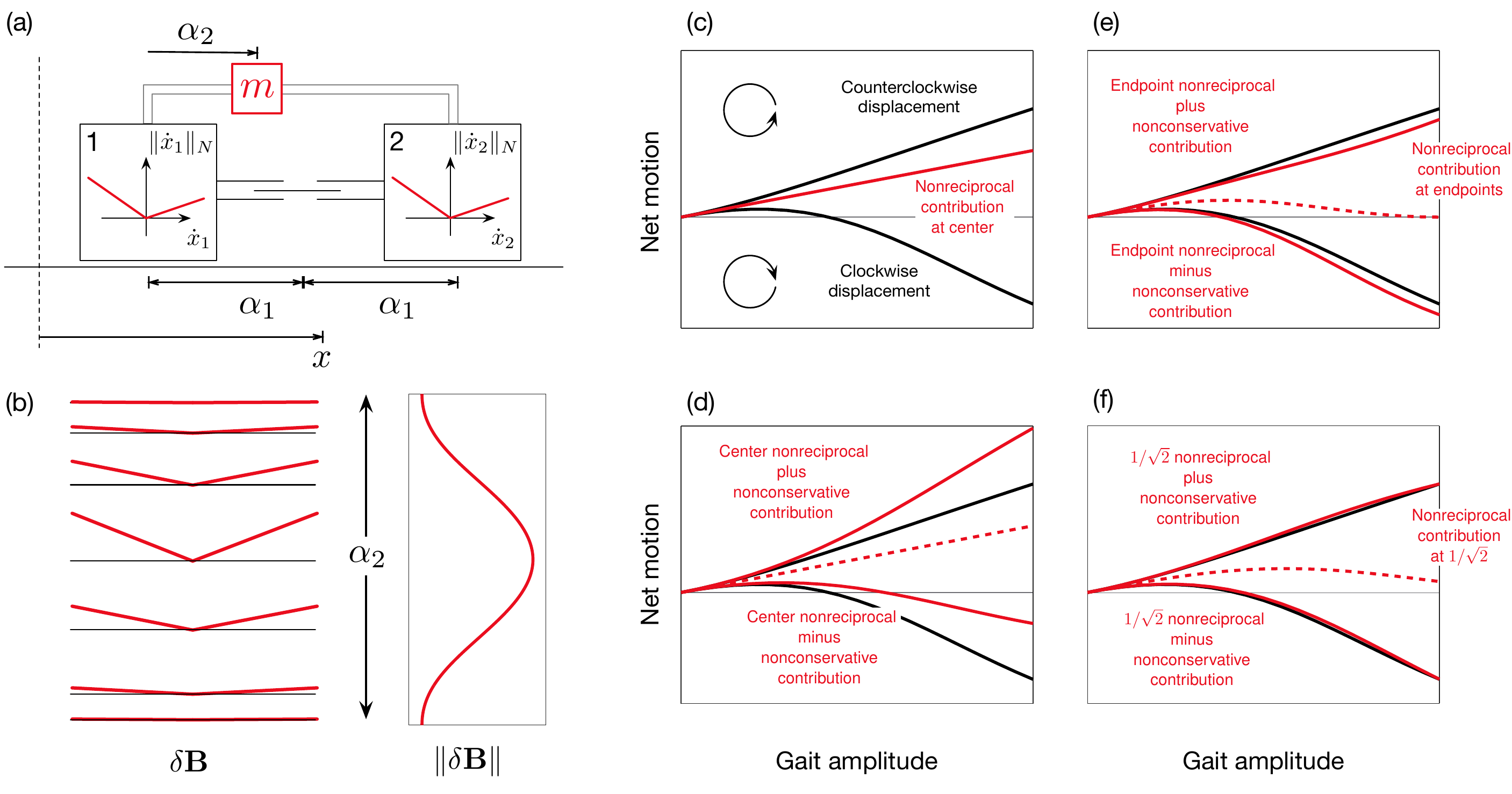}
\caption{Locomotion with nonconservative and nonreciprocal effects. (a) A two-block system with both the weight-shifting feature of the planarized skier of our early examples and the asymmetric friction from the later examples. (b) In addition to the same nonconservative curvature observed in the original planarized skier (see \zcref[S]{fig:riemanngeomech}) this skier has nonreciprocal curvature $\nonrecip\finconn=\finconndiff$ that is greatest when the weight is balanced between the blocks, and becomes weaker as the weight is concentrated over a single block. (c) The net motion induced by gait cycles of different size initially scale linearly in gait amplitude as per the nonreciprocal rules, but later diverge as the quadratically scaling nonconservative contributions become significant. The linear contribution is independent of the cyclic orientation of the gait, so gaits with either orientation scale along the same initial slope. (d) If the nonreciprocal curvature was constant, the net displacements would be a sum of the nonreciprocal slope plus or minus (depending on the gait orientation) a subquadratic nonconservative contribution. (e) Because $\nonrecip\finconn$ becomes smaller as the weight moves away from the center, the nonreciprocal contribution scales sublinearly with the amplitude of the gait. Evaluating the nonreciprocal curvature at the extremes of the gait (then multiplying it by the diameter of the gait) produces a much better estimate of the net displacement. (f) An even better estimate of the net displacement over an ellipse-shaped gait can be achieved by evaluating $\nonrecip\finconn$ at $1/\sqrt{2}$ of the $\jointangle_{2}$ amplitude, corresponding to the property that the $\jointangle_{1}$-direction motion during the gait happens when $\jointangle_{2}$ is near, but not at, its extreme value.}
\label{fig:asymskier}
\end{figure*}

\begin{example}
A cross-country skier with both  asymmetric friction on the skis and a weight-shifting shape mode, as illustrated in \zcref[S]{fig:asymskier}(a), can generate net motion from both nonreciprocal and nonconservative effects over gait cycles. Because the sign of the nonconservative contribution depends on the orientation of the gait cycle but the sign of the nonreciprocal contribution does not, the two effects either stack together or oppose each other, enabling fast motion in a preferred direction and slow motion in the other:

Using the same weight-shifting model of skiing as in the examples from \S\ref{sec:geomechoverview} (in which the drag coefficient on each foot is scaled by the square of the weight proportion on that foot), the $\finconnmean$ linear portion of the motility map and its $\extd \finconnmean$ nonconservative curvature are qualitatively the same as the motility map $\mixedconn$ illustrated in~\zcref[S]{fig:riemanngeomech}. The nonconservative contribution over a gait cycle thus scales subquadratically with the size of the gait, with the orientation of the cycle determining the sign of the resulting motion, with counterclockwise (positive) cycles contributing forward motion, and clockwise (negative) cycles contributing backwards motion.

The asymmetric friction on the blocks contributes a $\nonrecip \finconn$ nonreciprocal curvature term that is largest when the weight is centered over the two blocks, but diminishes as the weight shifts onto either block, as illustrated in \zcref[S]{fig:asymskier}(b).

Unlike the nonconservative contribution, the sign of the nonreciprocal contribution is independent of the orientation of the gait, and here is positive (or zero in the extremes of $\jointangle_{1}$ not moving or only moving when the weight is directly over a block). As illustrated in \zcref[S]{fig:asymskier}(c), the net motion over circular gaits at different amplitudes initially scales linearly with the amplitude of the gait, with the rate equal to $\nonrecip\finconn$ at the center of the gait.
As the quadratic growth of the nonconservative contributions catches up to the linear growth of the nonreciprocal contributions, the net motions diverge from this line, with the negative-orientation gaits eventually able to produce net backwards motion.

If the nonreciprocal component was independent of the position of the weight, the net motions would form symmetric subquadratic curves on either side of the nonreciprocal slope, as illustrated in \zcref[S]{fig:asymskier}(d). Because the $\nonrecip\finconn$ term grows smaller as the amplitude of the weight's motion increases, however, the actual nonreciprocal contribution is better described by one of the dashed curves in \zcref[S]{fig:asymskier}(e) or (f), which reflect the values of $\nonrecip\finconn$ at either the extremes of the weight amplitudes or at the extreme value scaled by $1/\sqrt{2}$ to account for some of the circular gait's $\jointangle_{1}$ motion happening while the weight is more balanced between the blocks.\footnote{If we had used a box-shaped gait for this example, all of the $\jointangle_{1}$ motion would have happened at the extreme of the $\jointangle_{2}$ range, and the endpoint value in (e) would have been a more accurate approximation than then scaled value in (f).} Applying these more accurate approximations of $\nonrecip\finconn$ pulls the net motions down, producing good approximations of the flat curve of positive-orientation displacements and the highly curved set of negative-orientation displacements.

\end{example}

\section{Conclusion}
\label{sec:conclusion}
Asymmetric friction is a fairly common phenomenon when contacts are made between a structured appendage and the environment.
In this paper we suggest that a broad swath of such asymmetric friction contacts can be effectively modeled by using a Finsler metric to express the power costs associated with the contact's sliding velocity.
In those systems in which friction annihilates momentum, this leads to a natural extension of principally kinematic motion from geometric mechanics.
In this extension a Finsler metric takes the place of the Reimannian metric that expresses dissipation.
The structure of the theory still leads to a motility map, but this map is no longer linear in the shape velocity $\jointangledot$---instead it is merely positively homogeneous and convex in $\jointangledot$.

We also identified a natural expansion for sub-Finslerian motility map, analogous to Taylor series polynomials.
The leading terms of this expansion are a linear form that captures the classical sub-Riemannian motility map, and a generalized cone whose principal components capture the primary ratcheting effects arising from the asymmetric friction on the system.
The linear form allows the suite of tools developed for geometric analysis of sub-Riemannian systems to be directly applied to the sub-Finslerian systems, and the cone part provides a geometric characterization of the primary differences in net motion that can be attributed to the asymmetry.

Future work will include the development of planning and gait optimization algorithms that utilize a Finsler metric framework, as well as the development of data-driven modeling tools which construct Finsler metric models from observation data.

A key target application of this future work will be to facilitate the design and use of robot feet.
Currently, most legged robots have ball feet or flat feet which can be modeled with linear drag friction models\footnote{
To the reader surprised by the notion that what is classically modeled as Coulomb friction contact can also be well-represented by linear drag, we suggest \cite[Sec. 4.2]{wu2025modeling}.%
}.
Yet directionally variable contact mechanisms appear in virtually every terrestrial clade of animals in the 1g to 10kg scale.
This evidence of convergent evolution, indicating a likely underlying functional advantage, suggests that the design space of feet, likely exploiting asymmetric friction forces, could have great practical importance.
Finsler metric friction models could be the key to the development of and planning for such novel robot feet, and our work here is a first step in exploring this new direction.

\appendix

\section{Convexity of Finsler metric level sets}\label{sec:apxA}

A key element of our approach is the property that aggregating the individual Finsler-drag terms for the individual pieces of the system produces a system-level Finsler metric, and in particular that the aggregate drag term has convex level sets.
In this appendix, we prove some key properties of positively homogeneous functions and convex functions that are used for showing these key results.

\subsection{Convexity Definitions and Lemmas}

\begin{definition}[Convex Combination]
Given an index set $I$, and maps $p: I \to V$ and $\alpha: T \to \mathbb{R}_{\geq 0}$, where $V$ is a real vector space.
If $\sum_{k\in I} \alpha(k) = 1$, then $\sum_{k\in I} \alpha(k) p(k) \in V$ is a ``Convex Combination'' of the points $p(I)$ with coefficients $\alpha(I)$.
More generally, the convex combination can be seen as a map $c: V^I \to V$ from any $I$-indexed set of points in any real vector space $V$ back into $V$, given by $c(s) := \sum_{k\in I} \alpha(k) p(k)$.
Note also that because the coefficients $\alpha(I)$ are non-negative, all the sums considered are absolutely convergent.
\end{definition}

\begin{definition}[Convex Set]
A set $S \subseteq V$ in a real vector space $V$ is ``convex'' if and only if every convex combination of elements of $S$ is in $S$.
\end{definition}

\begin{corollary}
In the simplest case, where the index set contains two elements: for all $x,y\in S$, $\alpha \in [0,1]$, we have $\alpha x + (1-\alpha) b \in S$.
\end{corollary}

\begin{definition}[Convex Function]\label{defn:convex}
A function $h: V \to \mathbb{R}$ from a real vector space $V$ into the reals is a ``Convex Function'' if for any convex combination $c$ and suitably indexed points $p$, it holds that $h(c(p)) \leq c(h(p))$.
Here we use the fact that the reals are themselves a vector space over the reals.
\end{definition}

\begin{definition}[Sub-level Set]
For $c \in \mathbb{R}$, and a function $h: V \to \mathbb{R}$ from a real vector space $V$ into the reals, the set $\{ x \in V | h(x) \leq c\}$ is the ``$c$ sub-level set of $h$''.
\end{definition}

\begin{lemma}The $v$~sub-level sets of a convex function are convex\label{lem:subLevelConvex}
\end{lemma}
\begin{proof}
Consider a convex combination $c$ indexed by $I$ with coefficients $\alpha$.
Assume that the points $p(k)$ for $k \in I$ belong to the $v$~sub-level set of a convex function $h: V \to \mathbb{R}$ from a real vector space $V$ into the reals.
We need to show that $h(c(p)) \leq v$, and thus $c(p)$ is in the $v$~sub-level set.
\begin{align}
h(c(p)) &= h\left(\sum_{k\in I} \alpha(k) p(k)\right) \leq
\sum_{k\in I} \alpha(k) h(p(k)) \nonumber\\
&\leq
\sum_{k\in I} \alpha(k) v \leq
v \sum_{k\in I} \alpha(k) = v.\qed
\end{align}
\end{proof}

\begin{lemma}\label{lem:convexInd}The convex functions on a real vector space $V$ taking only the values $0$ and $1$ are in a bijective correspondence with the convex subsets of $V$.
\end{lemma}
\begin{proof}
Consider a convex set $S \subseteq V$ in a real vector space $V$.
Define the ``convex indicator function'' $\bar\chi_S: V \to \mathbb{R}$ by $\bar\chi_S(v) = 1$ iff $v\notin S$, otherwise $\bar\chi_S(v) = 0$.

The function $\bar\chi_S(v)$ is (trivially) convex, giving one direction of the bijection.
The set $S$ is the fiber $\{ v \in V | \bar\chi_S(v) =0 \}$, giving the other direction of the bijection.  \qed
\end{proof}

\begin{lemma}Affine maps commute with convex combination\label{lem:affineConvex}
\end{lemma}
\begin{proof}
Consider an affine map $f: x\in  V \mapsto Ax+b \in U$ between real vector spaces $V$ and $U$.
For any convex combination $c$ indexed by $I$ with coefficients $\alpha$ and suitably indexed points $x$, we have
\begin{align}
c(f(x)) &=
\sum_{k\in I} \alpha(k) \left( A x(k) + b \right) \\ &=
\left(\sum_{k\in I} \alpha(k) A x(k)\right) + \left(\sum_{k\in I} \alpha(k) b \right) \\ &=
A \left(\sum_{k\in I} \alpha(k) x(k)\right) +  \left(\sum_{k\in I} \alpha(k) \right) b \\ &= A c(x) + b = f(c(x)).
\end{align}
\end{proof}

\begin{lemma}Convex Functions composed on Affine Maps are Convex Functions\label{lem:conCompAffine}
\end{lemma}
\begin{proof}
Consider an affine map $f: x\in  V \mapsto Ax+b \in U$ between real vector spaces $V$ and $U$, and a convex function $h: U \to \mathbb{R}$.
We wish to show that $h \circ f$ is a convex function on the real vector space $V$.

Consider a convex combination $c$ indexed by $I$ with coefficients $\alpha$, and suitably indexed points $x \in V$.
\begin{align}
(h \circ f \circ c)(x) = (h \circ c \circ f)(x) \leq (c \circ h \circ f)(x).
\end{align}
The equality follows from Lemma~\ref{lem:affineConvex}, the inequality from Definition~\ref{defn:convex}, and hence by Definition~\ref{defn:convex} itself, $h \circ f$ is a convex function.
\qed
\end{proof}

\begin{lemma}
Sums of convex functions are convex.\label{lem:convexSum}
\end{lemma}
\begin{proof}
Consider a real vector space $V$ with an index set $K$ indexing convex functions $h_k: V \to \mathbb{R}$ for $k\in K$.
Define the function $h: V  \to \mathbb{R}$ by $h(x) = \sum_{k\in K} h_k(x)$.
Consider also a convex combination $c$ indexed by $I$ with coefficients $\alpha$, and a compatibly indexed points $p$.
\begin{align}
h(c(p)) &= \sum_{k\in K} h_k\left( \sum_{i \in I} \alpha(i) p(i) \right) \nonumber\\
&\leq  \sum_{k\in K} \sum_{i \in I} \alpha(i) h_k(p(i))\nonumber\\
&= \sum_{i \in I} \alpha(i) \sum_{k\in K} h_k(p(i)) =c(h(p))
\end{align}

The final two double sums must agree because they are absolutely convergent, and we therefore have $h(c(p)) \leq c(h(p))$. \qed
\end{proof}

\newcommand{\bsHomog}{$\beta S$-Homogeneous}
\subsection{\bsHomog\ Definitions and Lemmas}

\begin{definition}[\bsHomog]
For $S \subseteq \mathbb{R}$, and $\beta: S \to \mathbb{R}$, a function $h: V \to U$ between two real vector space $V$ into is ``\bsHomog'' if and only if  for all choices of $v\in V$, $\alpha \in S$ it holds that $h(\alpha v) = \beta(\alpha) h(v)$.
\end{definition}

\begin{lemma}Finite Sums of \bsHomog\ functions are \bsHomog\label{lem:sumPosHom}
\end{lemma}
\begin{proof}
Given a real vector spaces $V$ and $U$, an index set $I$, a set of \bsHomog\
functions $h_k: V \to U$, $k\in I$, define the function $h(x) := \sum_{k\in I} h_k( x )$.
To examine whether $h(\cdot)$ is \bsHomog, consider any $\alpha \in S$:
\begin{align}
h(\alpha x) &= \sum_{k\in I} h_k( \alpha x ) =
\sum_{k\in I} \beta(\alpha) h_k( x ) \nonumber\\&=
\beta(\alpha) \sum_{k\in I} h_k( x ) = \beta(\alpha) h(x)
\end{align}
wherever the sums satisfy the distributive property.

In particular, the set of positively homogeneous functions between $V$ and $U$ is closed under finite sums. \qed
\end{proof}

\begin{lemma}\bsHomog\ functions of Linear Maps are \bsHomog\label{lem:posHomOfLin}
\end{lemma}
\begin{proof}
Given real vector spaces $V$,$U$ and $W$, a linear map $L:V \to U$, and a \bsHomog\ function $h: U \to W$, and any $\alpha \in S$ and $x \in V$,
\begin{align}
h(L(\alpha x)) = h(\alpha L(x)) = \beta(\alpha) h(L(x)).
\end{align}
It follows that $h \circ L$ is a \bsHomog\ map from $V$ to $W$.\qed
\end{proof}

\subsection{Positive Homogeneity Definitions and Lemmas}

Choosing $S$ to be the non-negative reals, and $\beta$ to be the identity map, \bsHomog\ restricts to the familiar concept of positive homogeneity.

\begin{definition}[Star Shaped Set]
A set $S \subseteq V$ in a real vector space $V$ is ``Star Shaped'' if for all $s\in S$ and $\alpha \in [0,1]$ it holds that $\alpha s \in S$.
\end{definition}

\begin{lemma}\label{lem:sls2ph}
The sub-level sets of any positively homogeneous function are star-shaped.
\end{lemma}
\begin{proof}
Let $s\in S_c$ for $S_c\subseteq V$ be the $c$ sub-level set of a positively homogeneous function $h: V \to \mathbb{R}$.
By definition of a sub-level set, $h(s) \leq c$.
Let $\alpha \in [0,1]$.
$h(\alpha s) = \alpha h(s) \leq \alpha c \leq c$, and therefore $\alpha s \in S_c$.\qed
\end{proof}

\begin{corollary}\label{coro:convexToStar}
All convex sets that contain $0$ are star-shaped.
\end{corollary}

\begin{lemma}
Star-shaped sets in a real vector space $V$ are in a bijective correspondence with the positively homogeneous functions on $V$.
\end{lemma}
\begin{proof}
We will construct a bijection between a positively homogeneous function $h: V \to \mathbb{R}$ and its $1$~sub-level set.

Lemma~\ref{lem:sls2ph} shows that the $1$-sub-level set of $h$ is a star shaped set, giving one direction of the bijection.

It remains to show that any star-shaped set is the $1$~sub-level set of a unique positively homogeneous function.
Consider a star-shaped set $S \subseteq V$.
For any $v \in V$, there exists a supremal $\beta \in \mathbb{R}\cup\{\infty\}$ such that $\beta v \in S$.
Define $h: V \to \mathbb{R}$ such that $h(v) := 1/\beta$.
For $\alpha\geq 0$ and the point $\alpha v$, that supremal value will therefore be $\beta / \alpha$.
It follows that $h(\alpha v) = 1/(\beta/\alpha) = \alpha h(v)$, i.e. $h$ is positively homogeneous.

It remains to show that $S$ is in fact the $1$~sub-level set of $h$, i.e. that if $s \in S$, then $h(s) \leq 1$, and if $h(s) \leq 1$ then $s\in S$.
If $s\in S$, then $1 s \in S$.
Since the $\beta$ value for this $s$ is the supremal multiplier with this property, it follows $\beta \geq 1$, and therefore $h(s) \leq 1$.

If $h(s) \leq 1$, then there exists a $\beta \geq 1$ such that $v := \beta s \in S$.
This implies that $s = (1/\beta) v$ where $1/\beta \in [0,1]$ and $v \in S$.
Because $S$ is star shaped, $s\in S$.\qed
\end{proof}

\section{Sub-Riemannian Approximation of Sub-Finslerian Motility Map} \label{app:subRapprox}

In this section we discuss computations that are all done at a constant shape $\jointangle$.
We therefore elide the $\jointangle$ parameter, and discuss e.g. $\finconn(\jointangledot)$ instead of $\finconn(\jointangle,\jointangledot)$.
Following \ref{eq:finconndecomposition} and \ref{eq:Btildedef} we write the approximation:
\begin{align}\label{eq:approxfcmd}
\finconn(\jointangledot) \approx \finconnmean \jointangledot + \finconndiff(\jointangledot)
= \finconnmean \jointangledot + \left(\jointangledot^T \mathbf{K}\jointangledot\right)^{\star\frac{1}{2}}.
\end{align}

We take the sub-Riemannian motility map $\finconnmean$ that best approximates $\finconn$ at a given shape
as being the linear map that minimizes the squared error between the velocities output by the sub-Riemannian and sub-Finslerian maps, on average for a uniform distribution on the unit sphere of shape velocities
\begin{align}
\finconnmean(\jointangle) := \arg\min_A \int\limits_{\|\jointangledot\|=1}\left\| A \jointangledot - \finconn(\jointangle,\jointangledot) \right\|^2~d\jointangledot.
\end{align}

Because $\finconn$ is positively homogeneous, this is the same as minimizing the expected squared relative error over a zero mean, direction independent gaussian shape velocity distribution, i.e.
\begin{align}
\finconnmean(\jointangle) &= \arg\min_A \mathbb{E}\left[ \frac{\left\| A \jointangledot - \finconn(\jointangle,\jointangledot) \right\|^2}{\|\jointangledot\|^2}\right] \nonumber\\
\jointangledot &\sim \mathrm{N}(0,\sigma^2).
\end{align}
Because $\finconn(\jointangle,\jointangledot)$ can be any convex positively homogeneous function of $\jointangledot$, it is an infinite dimensional object, and it is difficult to provide bounds on how accurately it can be estimated from any given finite sample of $\jointangledot$ values.
Additionally, the choice of norm used here weights the output of the approximation. For the sake of simplicity, all examples in this paper are calculated using the Euclidean norm in the working shape coordinates.

\subsection{Estimating the best linear part}
We estimate $\finconnmean(\jointangle)$ by generating a large sample of unit-norm shape velocity vectors $\hat{\jointangledot}_{k}$ evenly spaced around the unit circle.

We then use ordinary least-squares regression to estimate $\hat{\finconnmean}(\jointangle)$ as the linear map producing $\finconn(\jointangle, \hat{\jointangledot}_k)$ from $\hat{\jointangledot}_k$, setting up
\begin{equation}
\begin{bmatrix} \finconnmean_{1} & \finconnmean_{2} \end{bmatrix} \begin{bmatrix} \begin{matrix} \jointangledot_{1}^{1} \\[1ex]  \jointangledot_{1}^{2} \end{matrix} & \cdots & \begin{matrix} \jointangledot_{n}^{1} \\[1ex]  \jointangledot_{n}^{2} \end{matrix} \end{bmatrix} = \begin{bmatrix} \finconn(\jointangledot_{1}) &\ldots &\finconn(\jointangledot_{n}) \end{bmatrix}
\end{equation}
as the expected relationship if $\finconn$ were entirely linear, and then solving for the best linear fit by pseudoinverting the matrix of vector values as
\begin{equation}
\begin{bmatrix} \hat{\finconnmean}_{1} & \hat{\finconnmean}_{2} \end{bmatrix}  = \begin{bmatrix} \finconn(\jointangledot_{1}) &\ldots &\finconn(\jointangledot_{n}) \end{bmatrix} \pinv{\begin{bmatrix} \begin{matrix} \jointangledot_{1}^{1} \\[1ex]  \jointangledot_{1}^{2} \end{matrix} & \cdots & \begin{matrix} \jointangledot_{n}^{1} \\[1ex]  \jointangledot_{n}^{2} \end{matrix} \end{bmatrix}}.
\end{equation}

\subsection{Estimating the best conical part}
We employ a similar regression to find the generalized cone $\finconndiff$, through estimation of the matrix $\mathbf{K}$.
First note that squaring both sides of  \eqref{eq:approxfcmd} gives us a bilinear relationship between the residual value of $\finconn$ after removing $\finconnmean$ and the conical component of the motility map,
\begin{align}
\nonrecip\finconn(\jointangledot)^{\star2} = \left(\finconn(\jointangledot) - \finconnmean \jointangledot\right)^{\star 2}
\approx \finconndiff(\jointangledot)^{\star2} = \jointangledot^T \mathbf{K}\jointangledot.
\end{align}
We again use a standard least-squares regression to solve for the coefficients in $\mathbf{K}$, setting up
\begin{equation}
\dsf
\begin{bmatrix}
(\jointangledot_{1}^{1})^{2} &  \jointangledot_{1}^{1} \jointangledot_{1}^{2} & (\jointangledot_{1}^{2})^{2}
\\
\vdots & \vdots& \vdots
\\[1ex]
(\jointangledot_{n}^{1})^{2} &  \jointangledot_{n}^{1} \jointangledot_{n}^{2} & (\jointangledot_{n}^{2})^{2}
\end{bmatrix} \begin{bmatrix} \frac{1}{2} \mathbf{K}_{1} \\[1ex] \mathbf{K}_{12} \\[1ex] \frac{1}{2} \mathbf{K}_{2} \end{bmatrix} = \begin{bmatrix} \nonrecip\finconn(\jointangledot_{1})^{\star2} \\\vdots \\\nonrecip\finconn(\jointangledot_{n})^{\star2} \end{bmatrix}
\end{equation}
as the expected relationship if $(\nonrecip\finconn)^{\star2}$ was entirely a bilinear function, and then solving for a best linear fit by pseudoinversion as
\begin{equation}
\dsf
\begin{bmatrix} \frac{1}{2} \mathbf{K}_{1} \\[1ex] \mathbf{K}_{12} \\[1ex] \frac{1}{2} \mathbf{K}_{2} \end{bmatrix} = \pinv{    \begin{bmatrix}
(\jointangledot_{1}^{1})^{2} &  \jointangledot_{1}^{1} \jointangledot_{1}^{2} & (\jointangledot_{1}^{2})^{2}
\\
\vdots & \vdots& \vdots
\\[1ex]
(\jointangledot_{n}^{1})^{2} &  \jointangledot_{n}^{1} \jointangledot_{n}^{2} & (\jointangledot_{n}^{2})^{2}
\end{bmatrix}} \begin{bmatrix} \nonrecip\finconn(\jointangledot_{1})^{\star2} \\\vdots \\\nonrecip\finconn(\jointangledot_{n})^{\star2} \end{bmatrix}.
\end{equation}

\section{Baker-Campbell-Hausdorff Series with Asymmetry}
\label{app:BCH}

As an example of how asymmetric drag propagates through the Baker-Campbell-Hausdorff approximation for the net motion induced by a gait, we consider the special case of a box-shaped cycle whose sides are $abcd$ in cyclic order and have side lengths $L$ for  $a$ and $c$, and $\ell$ for $b$ and $d$ in shape space.
Each side has a corresponding averaged rigid body velocity $\grouplog{a}$, $\grouplog{b}$, $\grouplog{c}$, and $\grouplog{d}$, where e.g. $\grouplog{a}$ is the logarithm (in the Lie algebra sense) of the final point along $a$ relative to the initial point of $a$.
This closely follows the approach used in \citet{Bass:2022wn} eqns. (15) to (19), except here we consider a rectangular gait rather than a circular one.

We take motility map as having a value $\finconn \approx \finconnmean + \finconndiff$ at the center of the cycle and a first derivative $\frac{\partial\finconn}{\partial \jointangle} \approx \frac{\partial\finconnmean}{\partial\jointangle}+ \frac{\partial\finconndiff}{\partial\jointangle}$ around this point.
Further, we assume the sides $a$ and $c$ align with one principal axis of the cone $\finconndiff$ and the motion in the direction of $\jointangle_1$.
Consequently, $b$ and $d$ align with the other principal axis of the cone $\finconndiff$ and motion along the coordinate $\jointangle_2$.
Under these conditions, the integrated velocity over the four segments of the gait can be approximated as
\begin{subequations}
\begin{align}
\grouplog{a} &\approx \pmi \left(\finconnmean_{1} - \frac{\ell}{2}\frac{\partial \finconnmean_{1}}{\partial \jointangle_{2}} + \finconndiff_{1} - \frac{\ell}{2}\frac{\partial \finconndiff_{1}}{\partial \jointangle_{2}}\right)L
\\
\grouplog{b} &\approx \pmi \left(\finconnmean_{2} + \frac{L}{2}\frac{\partial \finconnmean_{2}}{\partial \jointangle_{1}} + \finconndiff_{2} + \frac{L}{2}\frac{\partial \finconndiff_{2}}{\partial \jointangle_{1}}\right) \ell
\\
\grouplog{c} &\approx -\left(\finconnmean_{1} + \frac{\ell}{2}\frac{\partial \finconnmean_{1}}{\partial \jointangle_{2}} - \finconndiff_{1} - \frac{\ell}{2}\frac{\partial \finconndiff_{1}}{\partial \jointangle_{2}}\right) L
\\
\grouplog{d} &\approx -\left(\finconnmean_{2} - \frac{L}{2}\frac{\partial \finconnmean_{2}}{\partial \jointangle_{1}} - \finconndiff_{2} + \frac{L}{2}\frac{\partial \finconndiff_{2}}{\partial \jointangle_{1}}\right) \ell.
\end{align}
\end{subequations}
(where we drop all derivatives of motility map components along their own directions, because they directly cancel to zero at this order of approximation).

Bringing the leading negative signs in $c$ and $d$ inside the parentheses as
\begin{subequations}
\begin{align}
\grouplog{a} &\approx \left(\pmi\finconnmean_{1} - \frac{\ell}{2}\frac{\partial \finconnmean_{1}}{\partial \jointangle_{2}} + \finconndiff_{1} - \frac{\ell}{2}\frac{\partial \finconndiff_{1}}{\partial \jointangle_{2}}\right)L
\\
\grouplog{b} &\approx \left(\pmi\finconnmean_{2} + \frac{L}{2}\frac{\partial \finconnmean_{2}}{\partial \jointangle_{1}} + \finconndiff_{2} + \frac{L}{2}\frac{\partial \finconndiff_{2}}{\partial \jointangle_{1}}\right)\ell
\\
\grouplog{c} &\approx \left(-\finconnmean_{1} - \frac{\ell}{2}\frac{\partial \finconnmean_{1}}{\partial \jointangle_{2}} + \finconndiff_{1} + \frac{\ell}{2}\frac{\partial \finconndiff_{1}}{\partial \jointangle_{2}}\right)L
\\
\grouplog{d} &\approx \left(-\finconnmean_{2} + \frac{L}{2}\frac{\partial \finconnmean_{2}}{\partial \jointangle_{1}} + \finconndiff_{2} - \frac{L}{2}\frac{\partial \finconndiff_{2}}{\partial \jointangle_{1}}\right)\ell
\end{align}
\end{subequations}
makes it straightforward to evaluate the lowest order terms of the BCH series.
To first order, the motion caused by the shape change $abcd$ is governed by the first order terms of the BCH series for that product:
\begin{subequations}
\begin{align}
&\grouplog{a}+ \grouplog{b}+ \grouplog{c}+ \grouplog{d}\\
&~~+ \frac12 \Bigl(\liebracket{\grouplog{a}}{\grouplog{b}} + \liebracket{\grouplog{a}}{\grouplog{c}}
+ \liebracket{\grouplog{a}}{\grouplog{d}}
+ \liebracket{\grouplog{b}}{\grouplog{c}}
+ \liebracket{\grouplog{b}}{\grouplog{d}}
+ \liebracket{\grouplog{c}}{\grouplog{d}} \Bigr)
\nonumber\\
&= \grouplog{a}+ \grouplog{b}+ \grouplog{c}+ \grouplog{d}\nonumber\\
&~~+ \frac{1}{2} \Bigl(\liebracket{\grouplog{a}}{(\grouplog{b}+\grouplog{c}+\grouplog{d})} + \liebracket{\grouplog{b}}{(\grouplog{c}+\grouplog{d})} + \liebracket{\grouplog{c}}{\grouplog{d}} \Bigr) \nonumber\\
&\approx \mathbf{0} + L\ell\left(\frac{\partial \finconnmean_{2}}{\partial \jointangle_{1}}- \frac{\partial \finconnmean_{1}}{\partial \jointangle_{2}}\right) + 2L\finconndiff_{1} + 2\ell\finconndiff_{2}  \label{eqn:bch1}\\
&~~+ L\ell\liebracket{\finconnmean_{1}}{\finconnmean_{2}} + \frac{1}{2}\liebracket{L\finconnmean_{1}+\ell\finconnmean_{2}}{L\finconndiff_{1}+\ell\finconndiff_{2}} \label{eqn:bch2}\\
&~~~+ \mathcal{O}(L^2\ell) + \mathcal{O}(L\ell^2) \label{eqn:bch3}.
\end{align}
\end{subequations}
Here \eqref{eqn:bch1} approximates the leading sum, \eqref{eqn:bch2} represents the first order contribution of the non-commutativity of the group, and \eqref{eqn:bch3} bounds the BCH approximation error.

Collecting the lowest-order terms then gives us a series approximation for the time-normalized geometric mean velocity over the gait,
\begin{equation}
\begin{split}
\grouplog{f} &\approx 2(L\finconndiff_{1} + \ell\finconndiff_{2})
\\
&+ L\ell\left(\frac{\partial \finconnmean_{2}}{\partial \jointangle_{1}}- \frac{\partial \finconnmean_{1}}{\partial \jointangle_{2}} \right)
\\
&+ L\ell\left(\liebracket{\finconnmean_{1}}{\finconnmean_{2}}\right)
\\
&+\frac{1}{2} \liebracket{(L\finconnmean_{1}+\ell\finconnmean_{2})}{(L\finconndiff_{1}+\ell\finconndiff_{2})},
\end{split}
\end{equation}
whose first three rows respectively encode the new nonreciprocal contribution to the net displacement, the standard nonconservative contribution, and the standard noncommutative contribution.

The fourth row captures noncommutative interactions between the symmetric and antisymmetric locomotion dynamics. Unlike the other second-order terms in $\grouplog{f}$, this term's values depend on the starting phase of the gait (i.e., which edge is traversed first). For example, if we take the gait cycle defined by $bcda$, the first three rows remain the same, but the sign on $\finconnmean_{1}$ flips. This phase-dependence is similar to the phase dependence we previously noted in~\cite{Bass:2022wn} for third-order terms in the standard symmetric-friction case; with asymmetric friction, the phase-dependence moves into the second-order terms. We will examine this phase-dependence more closely in future work on systems that both have asymmetric friction and can rotate to produce noncommutative effects.

\section{Funding}
The author(s) disclosed receipt of the following financial support for the research, authorship, and/or publication of this article: This work was supported in part by the United States Office of Naval Research, under Grant N00014-23-1-21, NSF CPS 2038432, and the D. Dan and Betty Kahn Foundation Autonomous Systems Mega-Project.
The authors would like to thank the students of the  ``Hands On Robotics'' class for the spines in \zcref[S]{fig:spines}.

\bibliographystyle{plainnat}

\end{document}